\documentclass[lettersize,journal]{IEEEtran}
\usepackage{array}
\usepackage[caption=false,font=normalsize,labelfont=sf,textfont=sf]{subfig}
\usepackage{textcomp}
\usepackage{stfloats}
\usepackage{url}
\usepackage{verbatim}
\usepackage{graphicx}
\usepackage{mathtools}
\usepackage{amsthm}
\usepackage{listings}
\usepackage{graphicx}
\usepackage{xcolor}
\usepackage{amsmath}
\usepackage{amssymb}
\usepackage{mathtools}
\usepackage{amsthm}
\definecolor{skyblue}{HTML}{0071BC}
\usepackage{color}
\usepackage{xspace}
\usepackage{overpic} 
\usepackage{multirow}
\usepackage{multicol}
\usepackage{booktabs}
\usepackage{bbding}
\usepackage{subcaption}
\usepackage{xspace}
\usepackage{soul}
\usepackage{bbding}
\usepackage{pifont}
\usepackage{colortbl}
\usepackage{enumitem}
\usepackage{overpic}
\usepackage{booktabs}
\usepackage[table]{xcolor}
\usepackage{dsfont}
\usepackage{wrapfig}
\usepackage{colortbl}
\usepackage{cite}
\usepackage{pifont}
\usepackage{makecell}
\usepackage{algorithm}  
\usepackage{algorithmic}  
\definecolor{citecolor}{HTML}{0071BC}
\usepackage[pagebackref=false, breaklinks=true, letterpaper=true, colorlinks, citecolor=citecolor, bookmarks=false]{hyperref}

\makeatletter
\DeclareRobustCommand\onedot{\futurelet\@let@token\@onedot}
\def\@onedot{\ifx\@let@token.\else.\null\fi\xspace}
\def\eg {\emph{e.g}\onedot} 
\def\ie{\emph{i.e}\onedot} 
 
\def\etc{\emph{etc}\onedot} \def\vs{\emph{vs}\onedot}

\makeatother

\begin{document}

\title{Unveiling the Secret of AdaLN-Zero in \\ Diffusion Transformer}

\author{Jie~Zhu, Mingyu~Ding, Boqiang~Duan, Leye~Wang,~\IEEEmembership{Member,~IEEE} and~Jingdong Wang,~\IEEEmembership{Fellow,~IEEE}
\thanks{Jie~Zhu, Leye~Wang are with Key Lab of High Confidence Software Technologies (Peking University), Ministry of Education, China, and School of Computer Science, Peking University, China.}
\thanks{Mingyu~Ding is with UC Berkeley.}
\thanks{Boqiang Duan and Jingdong Wang are with Baidu, Beijing, China.}
\thanks{E-mail: zhujie@stu.pku.edu.cn, myding@berkeley.edu, duanboqiang@baidu.com, leyewang@pku.edu.cn, wangjingdong@outlook.com}
\thanks{Corresponding author: Leye Wang, Jingdong Wang}
\thanks{Manuscript received April 19, 2021; revised August 16, 2021.}}

\markboth{Journal of \LaTeX\ Class Files,~Vol.~14, No.~8, August~2021}%
{Shell \MakeLowercase{\textit{et al.}}: A Sample Article Using IEEEtran.cls for IEEE Journals}


\maketitle

\begin{abstract}
%
%
Diffusion transformer (DiT), a rapidly emerging architecture for image generation, has gained much attention.
However, despite ongoing efforts to improve its performance, the understanding of DiT remains superficial.
%
In this work, we delve into and investigate a critical conditioning mechanism within DiT, adaLN-Zero, which achieves superior performance compared to adaLN.
%
Our work studies three potential elements driving this performance, including an SE-like structure, zero-initialization, and a ``gradual'' update order, among which zero-initialization is proved to be the most influential. Building on this understanding, we propose an analysis-guided initialization strategy, termed adaLN-Gaussian, which serves both as an empirical validation of our analysis and as a practical initialization method that consistently improves optimization efficiency. On the other hand, inspired by the SE-like structure, we introduce an improved conditioning mechanism called SE-adaLN-Zero. Extensive experiments following DiT on four datasets, especially on ImageNet1K, demonstrate the effectiveness and generalization of adaLN-Gaussian and SE-adaLN-Zero. Beyond class-to-image generation, we also evaluate the generalization of the two improved methods on text-to-image generation.

\end{abstract}


\section{Introduction}

Diffusion transformer (DiT)~\cite{peebles2023scalable} has recently emerged as a powerful architecture for image synthesis, and has gained vast attention for its superior performance over UNet-based diffusion models~\cite{dhariwal2021diffusion, rombach2022high}. 
As DiT continues to drive breakthroughs in image generation, there is a growing interest in pushing its performance boundaries even further.
Current efforts could be roughly categorized into two categories: 
1) those incorporating advanced techniques~\cite{chu2024visionllama,ma2024sit,lu2024fit,tian2024u,zhu2024sd, yaofasterdit}, like VisionLLama~\cite{chu2024visionllama}, which introduces language model-based tricks such as RoPE2D~\cite{su2024roformer} and SwishGLU~\cite{shazeer2020glu}, to boost the performance; 
and 2) those leveraging stronger and more informative conditions~\cite{esser2024scaling,chen2023pixart,chen2024pixart-v2,ma2024exploring,li2024hunyuan}, such as PixArt-$\alpha$~\cite{chen2023pixart} that extends DiTs to the text-to-image realm to enable more exquisite image generation.

Despite these advances, our understanding of the mechanisms driving DiT's performance remains superficial.
One critical aspect that requires further investigation is \textit{adaLN-Zero}, an important conditioning mechanism that significantly enhances DiT's performance compared to the original \textit{adaLN} (20.02 \textit{vs.} 24.13 in FID).
Fully understanding the underlying mechanism of adaLN-Zero is essential and may provide deeper insights for further optimizing DiT, especially given the increasing prevalence of DiT in the field of diffusion generation~\cite{karras2022elucidating, dhariwal2021diffusion,  karras2024analyzing, zhumole}.

In this work, we \textit{uncover the mechanism behind adaLN-Zero's performance boost, providing key insights into DiT’s conditioning process.}
By studying the differences between adaLN-Zero and adaLN,  our analysis studies three elements that collectively contribute to the performance enhancement: 1) a Squeeze-and-Excitation-like (SE-like) structure~\cite{hu2018squeeze}, 2) zero-initialized value (a well-optimized location in the optimization space), and 3) a ``gradual'' update order of model weights.
The SE-like structure arises from introducing scaling element $\alpha$ and the latter two stem from adaLN-Zero's zero-initialization strategy for $\alpha$. 
%
By empirical experiments, we find that a good zero-initialized location \textit{itself} plays a more significant role among the three elements. 
We reveal that compared to other initialization, zero-initialization enables the weights that derive $\alpha$ to morphologically more closely approximate the well-trained distribution which resembles a Gaussian distribution.
Interestingly, we find all the weights of condition modulations in DiT's blocks gradually form Gaussian-like distributions as training progresses.

Based on these findings, we propose to replace adaLN-Zero by initializing the weights of each condition modulation with Gaussian distributions, which we call adaLN-Gaussian.
Additionally, inspired by our analysis of SE-like structure, we introduce an improved conditioning mechanism termed SE-adaLN-Zero. To validate the effectiveness and generalization of adaLN-Gaussian and SE-adaLN-Zero, we conduct comprehensive experiments following DiT on four datasets, especially on ImageNet1K~\cite{imagenet15russakovsky}, testing across different training durations, DiT variants, improving strategies, and DiT-based models. Beyond class-to-image generation, we also evaluate the generalization of these two methods on text-to-image generation. Our contributions can be summarized as follows:

$\bullet$ We study three key factors that collectively contribute to the superior performance of adaLN-Zero: an SE-like structure, a good zero-initialized value, and a gradual weight update order. Among them, we find that a good zero-initialized value plays the most pivotal role.

$\bullet$ Based on the analysis about distribution variation of condition modulation weights, we heuristically leverage Gaussian distributions to initialize each condition modulation, termed adaLN-Gaussian.

$\bullet$ Drawing on our analysis of SE-like structures, we additionally propose an enhanced conditioning mechanism called SE-adaLN-Zero.

$\bullet$ Extensive experiments following DiT on four datasets, especially on ImageNet1K across different settings and text-to-image experiments validate the effectiveness and generalization of adaLN-Gaussian and SE-adaLN-Zero.

\begin{figure}[t]
\centering
\vskip -0.2in
\includegraphics[width=\linewidth]{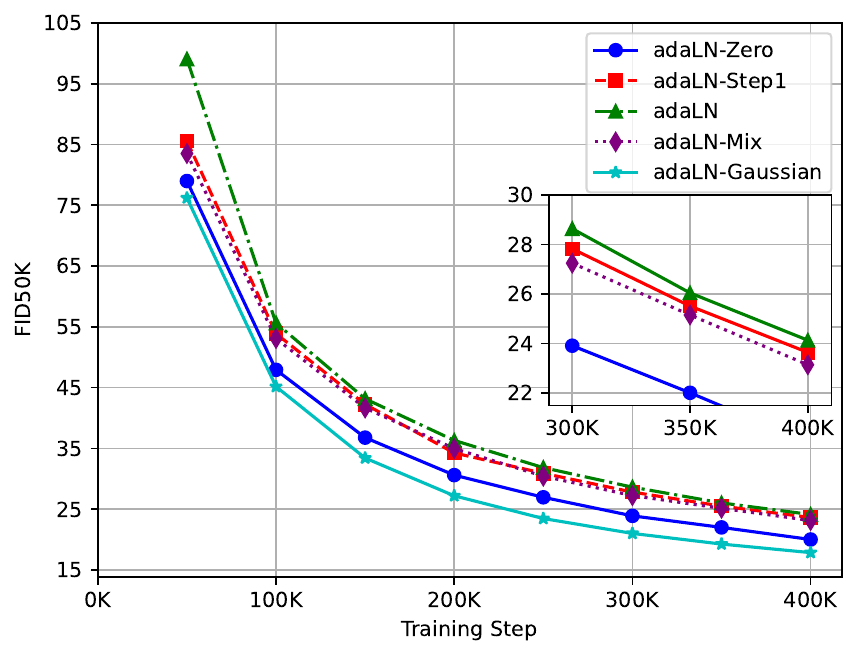}
        \caption{Comparing adaLN-Zero with adaLN, adaLN-Step1, adaLN-Mix, and adaLN-Gaussian on FID50K. We use the largest model DiT-XL/2 in all experiments on ImageNet1K $256\times256$.}
        \label{fig:fid}
        \vskip -0.1in
\end{figure}

\section{Related Work}

\textbf{Transformer in Diffusion.}  With the extensively demonstrated scalability and remarkable capabilities of transformers~\cite{vaswani2017attention, dosovitskiy2020image}, they have recently been introduced into diffusion generation~\cite{chai2023layoutdm, gao2023masked, mo2023dit, feng2023diffuser, feng2024latent, bao2023all, fei2024scaling,chen2024gentron,levi2023dlt,crowson2024scalable}. \cite{gao2023masked} propose an asymmetric masking diffusion transformer to explicitly enhance contextual relation learning among object semantic parts. DiffiT~\cite{hatamizadeh2023diffit} introduces hybrid hierarchical vision transformers with a U-shaped encoder and decoder. More recently, DiT~\cite{peebles2023scalable} replaces the widely-used UNet with transformers in diffusion generation, empirically demonstrating excellent performance and promising scalability. Subsequently, more efforts have been devoted to improving diffusion transformers. Following this research line, FiT~\cite{lu2024fit} and VisionLLama~\cite{chu2024visionllama} introduce large language model (LLM) techniques, such as RoPE2D~\cite{su2024roformer} and SwishGLU, to further enhance DiT. SD-DiT~\cite{zhu2024sd} incorporates masking operations into DiT to accelerate model convergence and improve performance. Pixart-$\alpha$ and Pixart-$\sigma$~\cite{chen2023pixart, chen2024pixart-v2} extend DiT to text-to-image synthesis and produces high-quality and exquisite images. U-DiT\cite{tian2024u} argues that the effectiveness of the U-Net inductive bias is meaningful but has been neglected in DiTs, reintroducing the U-shaped architecture to enhance performance. Additionally, SiT~\cite{ma2024sit} proposes a scalable interpolant framework built on the backbone of DiTs. FasterDiT~\cite{yaofasterdit} adopts advanced training and sampling strategies to improve performance without architecture modification. Different from these efforts, our work is motivated by elevating the understanding of DiT given its great prevalence in the generation realm, and focuses primarily on a crucial conditioning mechanism called adaLN-Zero. 

\textbf{Weight Initialization.} In a neural network, weight initialization is a crucial operation as it directly determines the initial position in the optimization space~\cite{narkhede2022review}. Typically, good initialization aids model training. Common methods include random initialization with (truncated) normal or uniform distributions. \cite{glorot2010understanding} introduced a properly scaled uniform distribution for initialization, known as ``Xavier" initialization, in \cite{jia2014caffe}. However, this strategy is not suitable for the ReLU activation function~\cite{nair2010rectified}, as ReLU can map negative values to zero, thereby altering the entire variance. To address this, \cite{he2015delving} proposed ``Kaiming" initialization, which assumes that half of the neurons are activated while the rest are zero. In the deep learning era, the zero-initialization strategy can be traced back to~\cite{goyal2017accurate}, where it was used to accelerate large-scale training 
potentially via nullifying certain output pathways to implicitly adjust the propagation of backward signals
in a supervised learning setting. 
More recently, it has been widely adopted in diffusion generation~\cite{ho2020denoising, rombach2022high} to ease optimization. In DiT~\cite{peebles2023scalable}, the impact of zero-initialization is particularly notable, leading to significant performance improvement. Motivated by this, we delve deeper into the underlying reasons, hoping that our findings will inspire further research.

\begin{figure}[t]
\centering
\vskip -0.2in
\includegraphics[width=\linewidth]{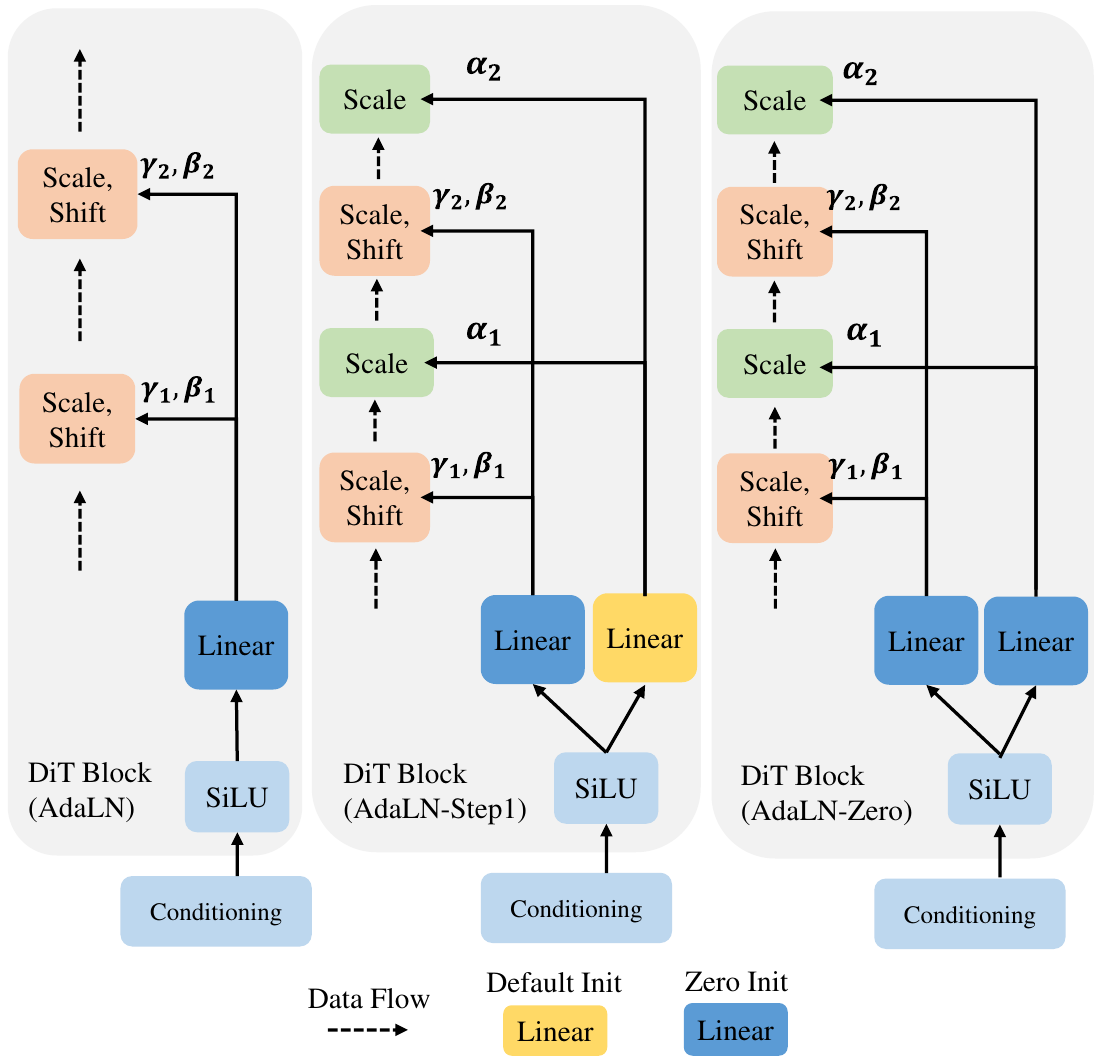}
\vskip 0.01in
        \caption{Illustration of adaLN, adaLN-Step1, and adaLN-Zero. The complete data flow in a DiT block is shown in Alg.~\ref{alg:Framwork}.}
        \label{fig:structure}
        \vskip -0.1in
\end{figure}

\section{Unveiling the Secret of AdaLN-Zero in DiT}

To unveil the underlying mechanism, we perform a detailed comparison between adaLN-Zero and adaLN. In Fig.~\ref{fig:structure} we find that adaLN-Zero introduces two additional steps: first, it introduces scaling element $\alpha$ (as denoted in DiT) for all transformer blocks; second, it zero-initializes corresponding linear layers to output zero vectors for all $\alpha$. Given these differences, one may naturally wonder: how do these two steps contribute to the performance gap between adaLN-Zero and adaLN in DiT?

\subsection{Decoupling adaLN-Zero by Evaluating Step One in Isolation} \label{sec:SE_analysis}

To answer this question, we decouple adaLN-Zero by introducing only the first step and initializing the linear layer's weights by default xavier uniform. For convenience, we denote this intermediate state as adaLN-Step1 as shown in Fig.~\ref{fig:structure} (middle). Then we train the three variants following the same training setting in DiT~\cite{peebles2023scalable} on ImageNet1K for 400K iterations using the largest and best-performing model, \ie, DiT-XL/2. Similarly, we measure FID~\cite{heusel2017gans} by
using ADM’s TensorFlow evaluation suite~\cite{dhariwal2021diffusion} following DiT and compare the performance of adaLN-Step1 with adaLN-Zero and adaLN in Fig.~\ref{fig:fid}. One can see that adaLN-Step1 outperforms adaLN even without zero-initializing the linear layer's weights, indicating that barely introducing scaling element $\alpha$ is beneficial as well. Similar results on Inception Score (IS)~\cite{salimans2016improved} could be found in \textcolor{red}{App.~\ref{app:is}}. Intuitively, adding scaling element $\alpha$ enhances adaLN's capability of expression, making model optimization easier and more flexible. Upon closer examination from overall structure, module function, and mathematical formula, we speculate that this improvement might be due to a Squeeze-and-Excitation-like (SE-like) architecture~\cite{hu2018squeeze}~\footnote{In \textcolor{red}{App.~\ref{App:se-structure}}, we provide the structure of SE module to better illustrate the similarity.}. Specifically, first, adaLN-Zero and SE module both serve as a side pathway compared to the main path. Second, scaling element $\alpha$ and SE module play a similar role, both of which aim to perform a channel-wise modulation operation. Third, formally, omitting the bias term, we illustrate the formulation of $\alpha$ in DiT in Eq.~\ref{eq:modulation} and SE module~\cite{hu2018squeeze} in Eq.~\ref{eq:SE}, respectively, with slight adjustments to make the two formulas more comparable:
\begin{equation} \label{eq:modulation}
  F(c)= \underbrace{(c \odot \operatorname{Sigmoid}(c))}_{\operatorname{SiLU}} * W_{\alpha} =  (c \odot (\operatorname{Sigmoid}(I * c))) * W_{\alpha} \,,
\end{equation}
\begin{equation} \label{eq:SE}
  SE(c) = (\operatorname{ReLU}(W_{1} * c)) * W_{2} = ( \mathds{1} \odot (\operatorname{ReLU}(W_{1} * c))) * W_{2} \,,
\end{equation}
where $*$ is matrix multiplication, $\odot$ is Hadamard product, and $\mathds{1}$ is a vector full of element $1$. To some extent, it is observed that $F(c)$ shares a similar formulation with $SE(c)$. Given that $SE(c)$ has been extensively demonstrated to enable a general enhancement over various vision tasks~\cite{hu2018squeeze}, this similarity may contribute to the improved performance of adaLN-Step1.

On the other hand, it is worth noting that while adaLN-Step1, \ie, the first step, does contribute positively, there remains a large performance disparity between adaLN-Zero and adaLN-Step1. This suggests that the zero-initialization strategy, \ie, the second step, is equally necessary. We explore this further in the next subsection for clarity.


\subsection{How Zero-initialization Improves the Performance}

For a typical initialization strategy, \eg, kaiming initialization~\cite{he2015delving}, its \textit{fundamental} role is to \textit{determine} the initial location of the model in the optimization space. Particularly, in the case of zero-initialization, besides this function, \cite{goyal2017accurate} suggest that it also has an \textit{additional} role. Specifically, it can implicitly adjust the model structure by nullifying certain output pathways at the beginning of training, more importantly, causing the forward/backward signals to initially propagate through the identity shortcut~\cite{he2016deep}, thereby easing the optimization at the start of training~\cite{goyal2017accurate}. However, is this additional role really responsible for the performance gap between adaLN-Zero and adaLN-Step1? To answer this question, we first examine how this additional role specifically impacts optimization through the lens of gradient update. Afterward, we decouple this impact on gradient update during training to highlight the fundamental role of zero-initialization.

\begin{algorithm}[t] 
\caption{Forward Process of DiT with One Block}  
\label{alg:Framwork}  
\begin{algorithmic}[1]  
    \REQUIRE Noise disturbed latent $x$; A simplified DiT: PatchEmbed $W_{pat}$, matrix $W_{\beta_{1}}$, $W_{\gamma_{1}}$, and $W_{\alpha_{1}}$ deriving $\beta_1$, $\gamma_1$, and $\alpha_1$ for the first modulation, a linear layer $W_{att}$ replacing self-attention, matrix $W_{\beta_{2}}$, $W_{\gamma_{2}}$, and $W_{\alpha_{2}}$ deriving $\beta_2$, $\gamma_2$, and $\alpha_2$ for the second modulation, a linear layer $W_{ffm}$ replacing pointwise feedforward, $\gamma_f$, $\beta_f$, and $W_{f}$ for the modulation and linear layer in FinalLayer, respectively;  
    \ENSURE  
    predicted $\bar{\epsilon}$;
    \STATE $x_p$ = $x$ * $W_{pat}$  \quad \textcolor{gray}{\# Reshape and patchify $x$}
        \STATE
        \STATE \textcolor{gray}{\# DiT Block}
        \STATE $x_{m_1}$ =  $x_p$ $\odot$ $(1+\gamma_1)$ + $\beta_1$ 
\quad \textcolor{gray}{\# Modulation}
        \STATE $x_{att}$ = $x_{m_1}$ * $W_{att}$ \quad \textcolor{gray}{\# Replace attention}
        \STATE $x_{out_1}$ = $x_{att}$ $ \odot $ $\alpha_1$ + $x_p$ \quad \textcolor{gray}{\# Skip connection}
        \STATE $x_{m_2}$ =  $x_{out_1}$ $ \odot $ $(1+\gamma_2)$ + $\beta_2$ \quad \textcolor{gray}{\# Modulation}       
        \STATE $x_{ffm}$ = $x_{m_2}$ * $W_{ffm}$ \quad \textcolor{gray}{\# Replace FFM}
         \STATE $x_{out_2}$ = $x_{ffm}$ $\odot$ $\alpha_2$ + $x_{out_1}$ \quad \textcolor{gray}{\# Skip connection}
         \STATE
         \STATE $x_{f}$ =  $x_{out_2}$ $ \odot $ $(1+\gamma_f)$ + $\beta_f$ 
\quad \textcolor{gray}{\# Modulation}
        \STATE $\bar{\epsilon}$ = $x_{f}$ * $W_f$
     \RETURN $\bar{\epsilon}$
\end{algorithmic}
\end{algorithm} 

\subsubsection{Zero-initialization's Impact on Gradient Update} \label{sec:answer1}

Considering the complexity of the DiT model, we make three reliable modifications to simplify our gradient derivation. First, we use only one DiT block, easing the computations of complex chain rules. Second, we replace the multi-head self-attention and pointwise feedforward modules within the DiT block with simple linear transformations, respectively. Though this replacement alters the structure of the DiT block, from the view of backpropagation it does not affect the gradient flow of other modules but itself which is not our emphasis. Therefore this adjustment could be acceptable. Finally, for a linear layer, we omit the bias term in both the forward and backward passes. These alterations significantly simplify our analysis without negatively impacting the conclusions. We formally present the mathematical forward process in Alg.~\ref{alg:Framwork}. Note that in DiT, LayerNorm is learning-free, so we omit it from our formulation. The process of gradient derivation for each module weight is provided in \textcolor{red}{App.~\ref{app:gradient dereviation}}.  

To continue our analysis, reviewing the initialization strategy of DiT is necessary. adaLN and adaLN-Zero both initialize the FinalLayer module to zero, indicating that $\gamma_f$ ($W_{\gamma_f}$), $\beta_f$ ($W_{\beta_f}$), and $W_f$ are all zero at the beginning. As shown in Fig.~\ref{fig:structure}, adaLN and adaLN-Zero also zeros out weights of all modulations including $W_{\gamma_1}$, $W_{\beta_1}$, $W_{\gamma_2}$, and $W_{\beta_2}$ in a block, rendering $\gamma_1$, $\beta_1$, $\gamma_2$, and $\beta_2$ zero. A key difference from adaLN is that  
adaLN-Zero not only introduces $W_{\alpha_1}$ and $W_{\alpha_2}$ to produce scale parameters $\alpha_1$ and $\alpha_2$ (\ie, adaLN-Step1) \textbf{\textit{but also zero out $W_{\alpha_1}$ and $W_{\alpha_2}$ to make $\alpha_1$ and $\alpha_2$ become zero.}} See Tab.~\ref{tab:update order} 2nd row.

Therefore, \textit{in this first forward pass}, $W_f = 0$ and output is zero  (\textcolor{red}{See App.~\ref{app:gradient dereviation} Eq.~2}). Interestingly, \textit{in the first backward pass}, the gradient of $W_f$, \ie, $\frac{\partial \mathcal{L}}{\partial W_{f}}$, is not zero while the gradients of the rest, \ie, $\frac{\partial \mathcal{L}}{\partial W_{ffm}}$, $\frac{\partial \mathcal{L}}{\partial W_{att}}$, $\frac{\partial \mathcal{L}}{\partial W_{pat}}$, $\frac{\partial \mathcal{L}}{\partial W_{\gamma_f}}$, $\frac{\partial \mathcal{L}}{\partial W_{\beta_f}}$, $\frac{\partial \mathcal{L}}{\partial W_{\alpha_2}}$, $\frac{\partial \mathcal{L}}{\partial W_{\gamma_2}}$, \etc, are zero as their gradient formulas all include $W_f$ term and $W_f = 0$. Hence, only $W_{f}$ is updated while the rest weights are kept. So how about the next? \textit{In the second backward pass}, though $W_f$ is not zero, \textbf{the zero-initialized $\alpha_1$ and $\alpha_2$ due to adaLN-zero cause $\frac{\partial \mathcal{L}}{\partial W_{ffm}}$, $\frac{\partial \mathcal{L}}{\partial W_{att}}$, $\frac{\partial \mathcal{L}}{\partial W_{\gamma_2}}$, 
$\frac{\partial \mathcal{L}}{\partial W_{\beta_2}}$, $\frac{\partial \mathcal{L}}{\partial W_{\gamma_2}}$, and 
$\frac{\partial \mathcal{L}}{\partial W_{\beta_2}}$ to remain zero.} How about the third iteration? To better illustrate the gradient variation of involved weights, we show the gradient of all weights in the first several iterations in Tab.~\ref{tab:update order}. One can see that all weights do not update together as expected but \textit{gradually} update. Specifically, in the 1st iteration, only $W_{f}$ updates. \textbf{\textit{In the $2$nd iteration, only $W_{f}$, $W_{pat}$, $W_{\gamma_f}$, $W_{\beta_f}$, $W_{\alpha_2}$, and $W_{\alpha_1}$ update, which is what zero-initialization brings to the optimization update.}} In other words, zero-initialization introduces an additional ``gradual" update in the initial stage of optimization compared to adaLN-Step1.

\setlength{\tabcolsep}{0.25cm}{\begin{table*}[t]
		\begin{center}
        \caption{Gradient of different weights during training. The first row is the state of parameters' initial weight. $0$ means that the weight is zero. \Checkmark means that the gradient is not zero and the weight effectively updates while \XSolidBrush means the gradient is still zero and the weight does not update.} 
        \label{tab:update order}
			\begin{tabular}{c c c c c c c c c c c c c}
				\toprule
				Time/Gradient & $\frac{\partial \mathcal{L}}{\partial W_{f}} $& $\frac{\partial \mathcal{L}}{\partial W_{ffm}}$ & $\frac{\partial \mathcal{L}}{\partial W_{att}}$ & $\frac{\partial \mathcal{L}}{\partial W_{pat}}$ & $\frac{\partial \mathcal{L}}{\partial W_{\gamma_f}}$ &$\frac{\partial \mathcal{L}}{\partial W_{\beta_f}}$ &$\frac{\partial \mathcal{L}}{\partial W_{\alpha_2}}$ & $\frac{\partial \mathcal{L}}{\partial W_{\gamma_2}}$ &$\frac{\partial \mathcal{L}}{\partial W_{\beta_2}}$ & $\frac{\partial \mathcal{L}}{\partial W_{\alpha_1}}$ & $\frac{\partial \mathcal{L}}{\partial W_{\gamma_1}}$ & $\frac{\partial \mathcal{L}}{\partial W_{\beta_1}}$ \\ 
        \midrule
    Initial weight & 0 & $W_{ffm}$ & $W_{att}$ &  $W_{pat}$ & 0 & 0& 0 &0 &0 & 0& 0 & 0 \\
				\midrule
                    $1$st iteration &  \Checkmark & \XSolidBrush & \XSolidBrush & \XSolidBrush & \XSolidBrush & \XSolidBrush & \XSolidBrush & \XSolidBrush & \XSolidBrush & \XSolidBrush & \XSolidBrush & \XSolidBrush \\
                     \midrule 
                     \rowcolor{gray!30}
                    $2$nd iteration &  \Checkmark & \XSolidBrush & \XSolidBrush & \Checkmark & \Checkmark & \Checkmark& \Checkmark & \XSolidBrush & \XSolidBrush & \Checkmark & \XSolidBrush & \XSolidBrush \\
                                         \midrule   
                    $3$rd iteration &  \Checkmark & \Checkmark & \Checkmark & \Checkmark & \Checkmark & \Checkmark& \Checkmark & \Checkmark & \Checkmark & \Checkmark& \Checkmark & \Checkmark \\
				\bottomrule
			\end{tabular}
		\end{center}
        \vskip -0.1in
\end{table*}}

\textbf{Remark.} It is worth noting that although our derivation is based on a simplified version of DiT, we corroborate that this update order aligns with that of original DiT variants in which for a typical DiT model (adaLN-Zero), $W_{f}$ update first, subsequently, $W_{pat}$, $W_{\gamma_f}$, $W_{\beta_f}$, and \textit{all} $W_{\alpha}$~\footnote{For brevity, we use $W_{\alpha}$ to denote all $W_{\alpha_{1}}$ and $W_{\alpha_{2}}$ in DiT's blocks. $W_{\gamma}$ and $W_{\beta}$ are the same.} can update, and finally all parameters start to update. This verification demonstrates that our simplification is reasonable and our derivation is right. 

\subsubsection{Decoupling the Impact of Zero-initialization} 

Based on Sec.~\ref{sec:answer1}, we know that beyond the difference of initial position in the optimization space, the additional distinction between adaLN-Zero and adaLN-Step1 lies in the \textit{second} iteration of gradient optimization, where adaLN-Zero preferentially optimizes $W_{f}$, $W_{pat}$, $W_{\gamma_f}$, $W_{\beta_f}$, and \textit{all} $W_{\alpha}$~\footnote{For adaLN-Step1 (as well as adaLN), the model begins updating all weights after the first iteration, unlike typical initialization strategies where all weights are updated from the very beginning. We will explore the impact of this difference in future work.} while adaLN-Step1 optimizes all weights. Considering the performance disparity between adaLN-Zero and adaLN-Step1, is this update discrepancy crucial for enhancing model performance? Or is it just the zero-initialized position in optimization space that contributes more?

\begin{figure}[t]
\centering
\includegraphics[width=0.9\linewidth]{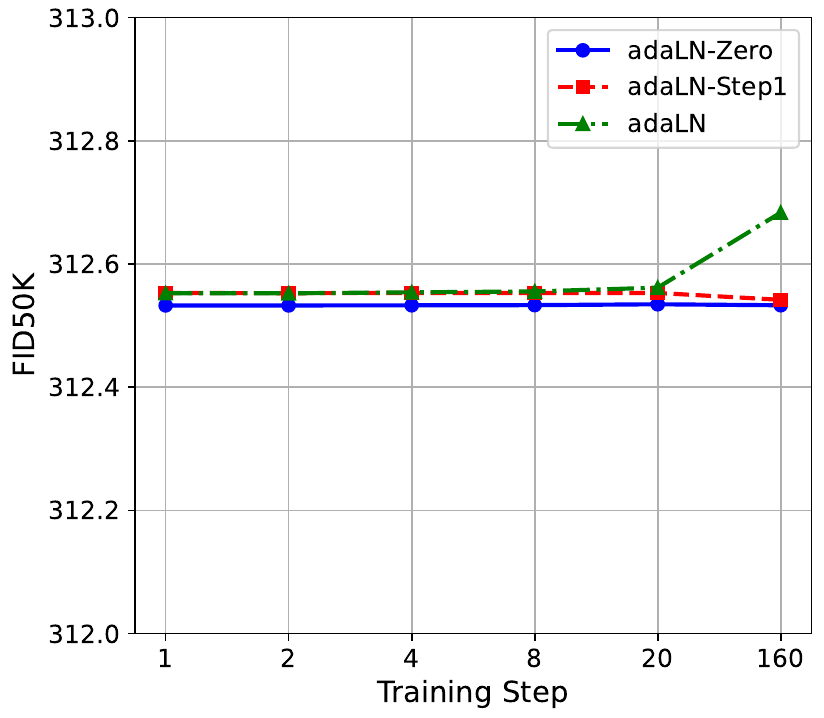}
    \caption{Performance of different variants in the initial stage.}
    \vskip -0.1in
    \label{fig:early-performance}
\end{figure}

Intuitively, if this update discrepancy is critical, we would see a \textit{significant} performance variation between adaLN-Zero and adaLN-Step1 within the initial few iterations since this discrepancy only occurs during the second iteration~\footnote{In our view, if the discrepancy in gradient updates is crucial, it can \textit{significantly} affect performance in the short term. And as the update period extends, the impact of this discrepancy diminishes.}.
Thus, we evaluate model performance during the early iterations, as shown in Fig.~\ref{fig:early-performance}. 
The results indicate minimal performance fluctuation between adaLN-Zero and adaLN-Step1 during the first 160 iterations, suggesting that the discrepancy in update order may not be as critical as initially expected.
Similar results can be found for the Inception Score (IS) in \textcolor{red}{App.~\ref{app:is}}.

To formally verify our hypothesis, we design an ingenious experiment to decouple the impact of zero-initialization on gradient update. Specifically, considering that the additional effect on the gradient cannot be avoided when zeroing out $W_{\alpha}$, we adopt the initialization of adaLN-Step1 but enforce the update order of adaLN-Zero simultaneously. We refer to this hybrid strategy as adaLN-Mix and compare its performance with adaLN-Zero and adaLN-Step1 in Fig.~\ref{fig:fid}. It is seen that while adaLN-Mix further enhances the performance of adaLN-Step1, it still lags significantly behind adaLN-Zero. This first indicates that the update order resulting from zero-initialization does contribute independently to performance. However, this contribution is not the primary reason for the substantial performance improvement seen in adaLN-Zero. In other words, it is the zero-initialized location that accounts for the remarkable performance difference between adaLN-Zero and adaLN-Mix. Similar results on Inception Score (IS) could be found in \textcolor{red}{App.~\ref{app:is}}. Why a zero-initialized location is such important, we put further exploration in the next subsection for clarity. 


\begin{figure*}[t]
\vskip -0.2in
\centering\centerline{\includegraphics[width=0.95\linewidth]{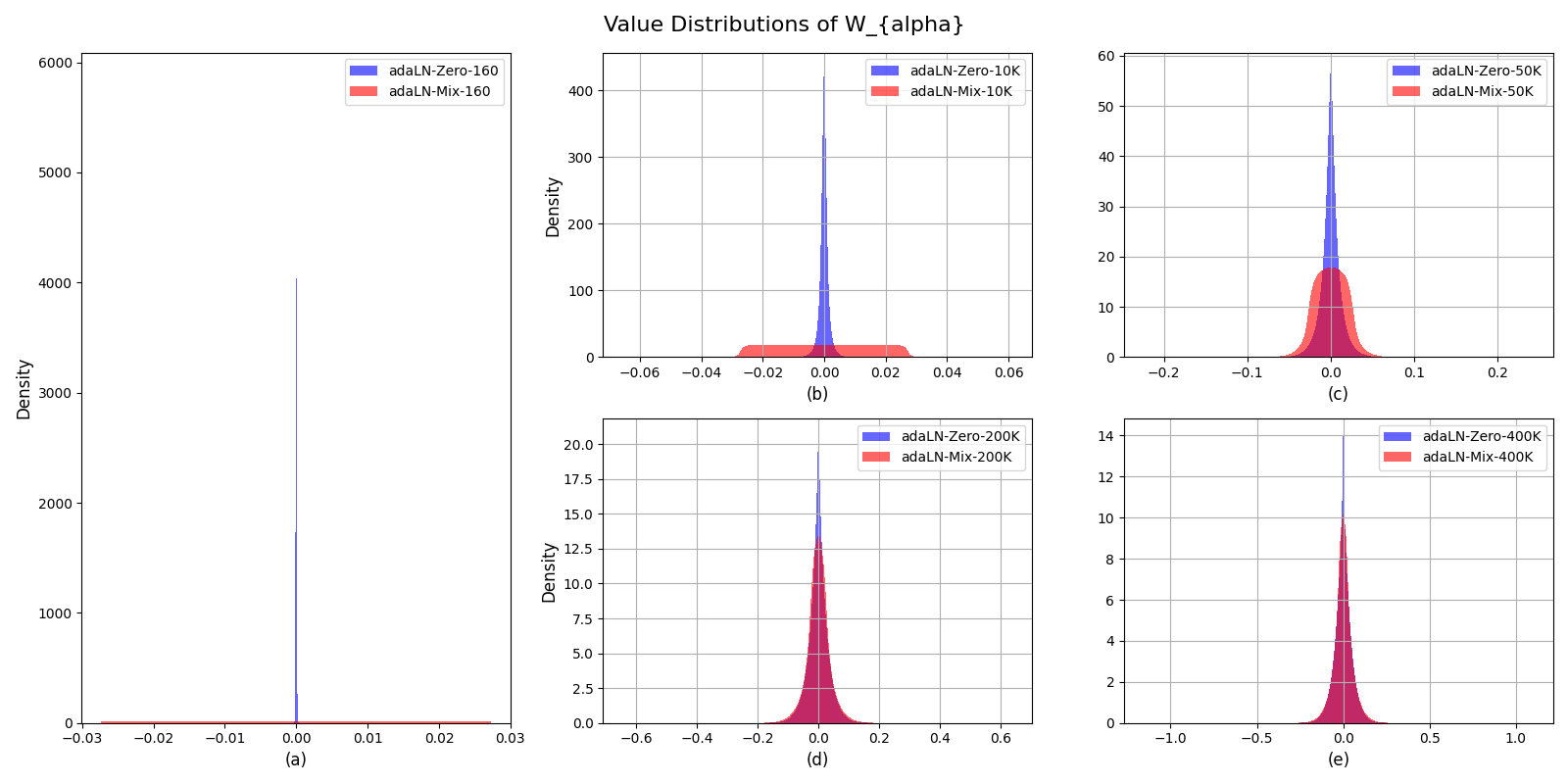}}
\caption{Value distributions of all $W_{\alpha}$ during the training process. AdaLN-Zero and adaLN-Mix are initialization strategies and 160, 10K, and 50K are timestamps.}
\label{fig:distribution}
\vskip -0.1in
\end{figure*}

\subsection{Why A Zero-initialized Location Wins?} \label{sec:win}

A simple answer might be that zero-initialization avoids introducing noise, as zero is a relatively neutral choice. However, this explanation is neither direct nor fully satisfying, so we aim to unveil a more fundamental reason. Our analysis begins by examining the variation in the weight distribution of all $W_{\alpha}$ in adaLN-Zero and adaLN-Mix~\footnote{We use adaLN-Mix instead of adaLN-Step1 to eliminate the potential influence of the discrepancy in update order of weights.}, respectively, as training progresses. 

\begin{figure}[t]
    \centering
    \includegraphics[width=0.9\linewidth]{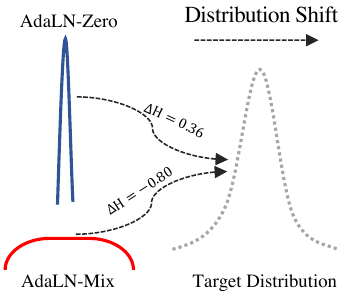}
    \caption{An abstract illustration of the entropy analysis on distribution movement for adaLN-Zero and adaLN-Mix.}
    \label{fig:abstract}
    \vskip -0.15in
\end{figure}

\begin{figure}[h]
\vskip -0.1in
\centering\centerline{\includegraphics[width=0.95\linewidth]{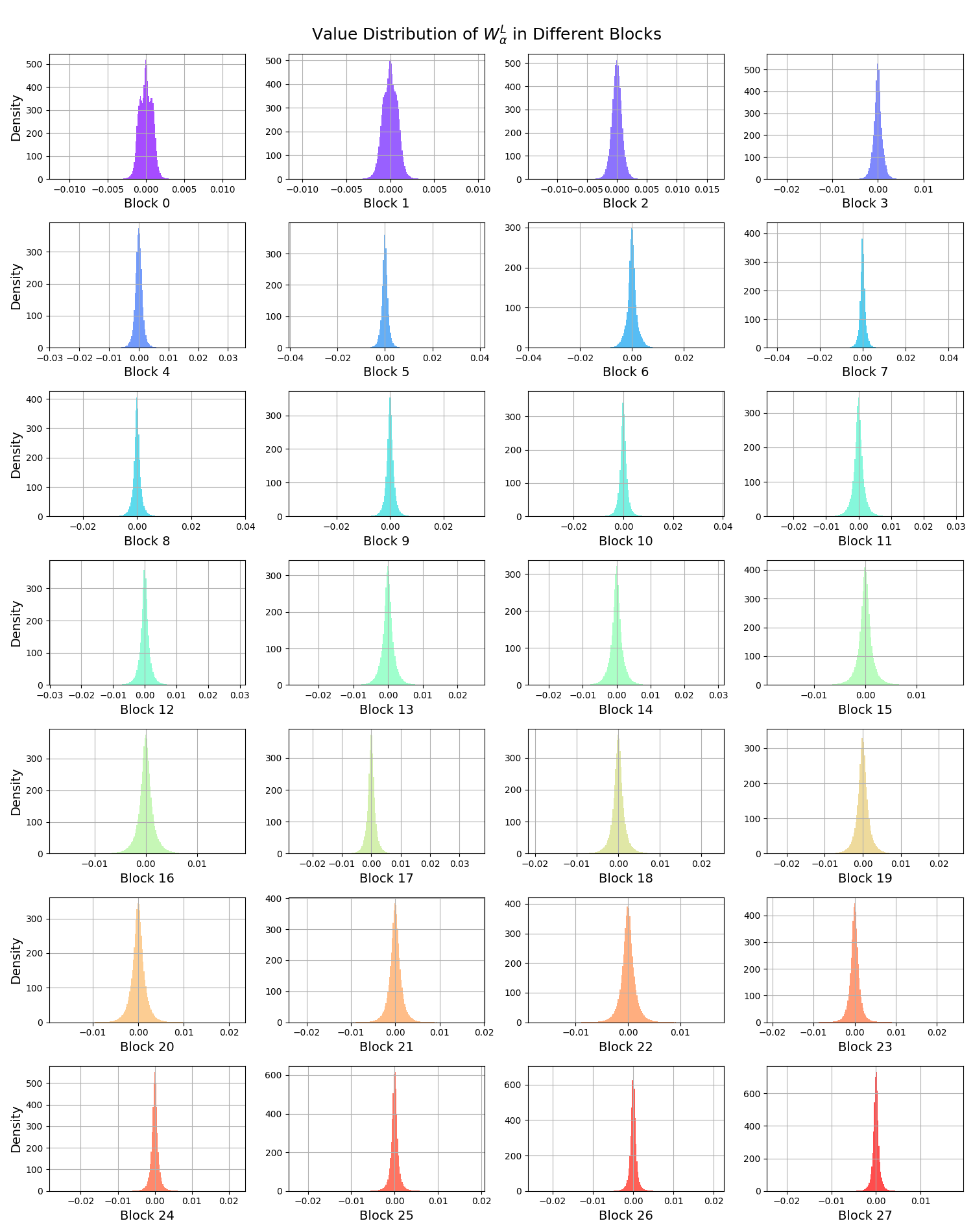}}
\caption{Value distributions of $W_{\alpha}^{L}$ in different blocks.}
\label{fig:each-block}
\vskip -0.1in
\end{figure}

\begin{figure*}[htbp]
    \centering
    \vskip -0.05in
    \includegraphics[width=1.0\linewidth]{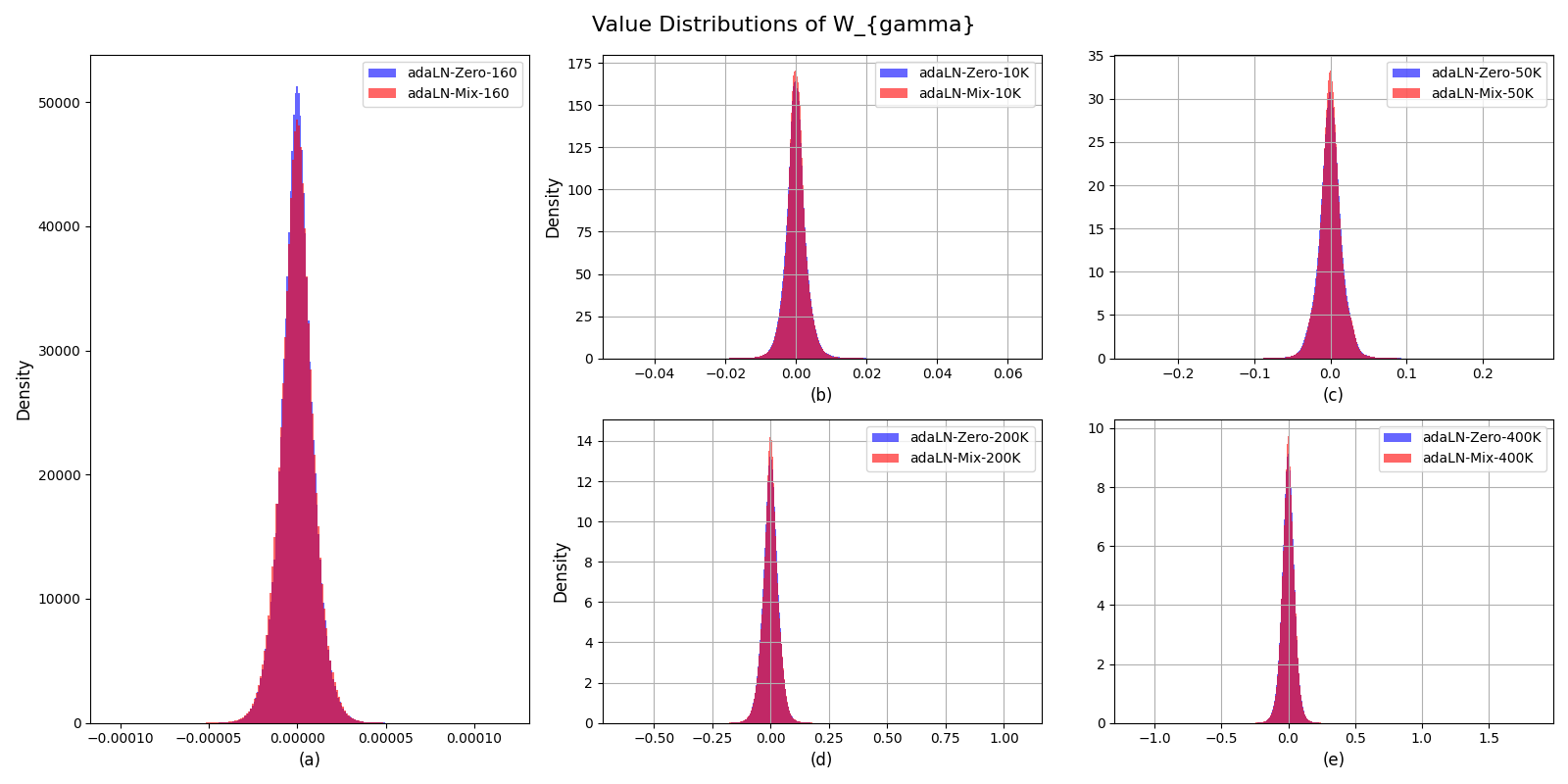}
    \vskip -0.05in
    \caption{Value distributions of the whole $W_{\gamma}$ in DiT blocks during the training process.}
    \label{fig:gamma}
    \vspace{1em} 
    \includegraphics[width=1.0\linewidth]{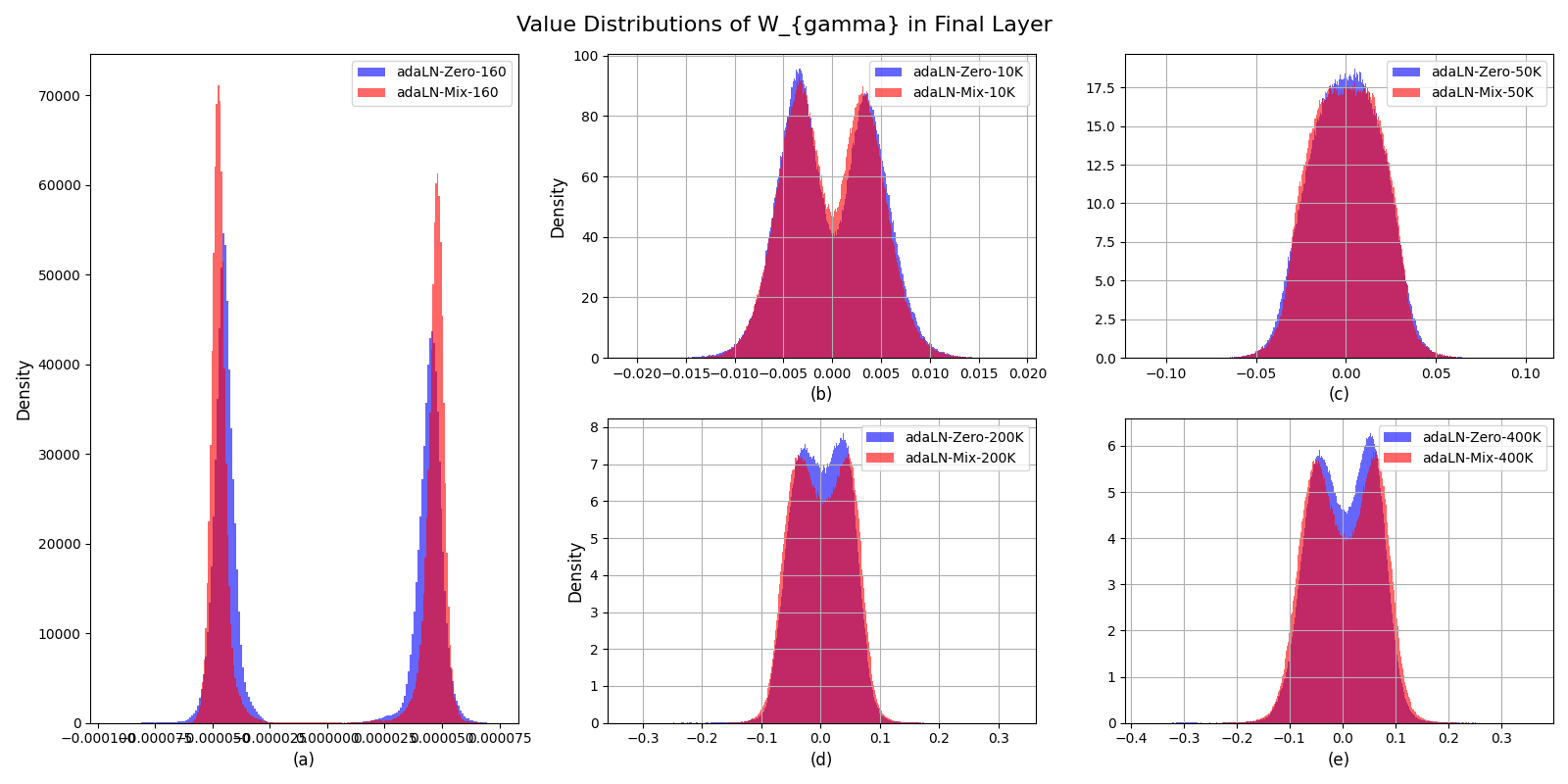}
    \vskip -0.05in
    \caption{Value distributions of $W_{\gamma_{f}}$ in FinalLayer during the training process.}
    \label{fig:gamma_f}
\end{figure*}



As illustrated in Fig.~\ref{fig:distribution}, we record the distribution of the entire $W_{\alpha}$ in 160, 10K, 50K, 200K, and 400K iterations, respectively, to observe the pattern of weight variation over time. At the start, as seen at 160 iterations in Fig.~\ref{fig:distribution} (a), adaLN-Zero exhibits a completely vertical distribution with most values being zero, while adaLN-Mix shows a completely horizontal distribution with a large span of value compared to adaLN-Zero, forming a nearly orthogonal relationship. As the training progresses, (\eg, from 160 to 400K), the distribution of adaLN-Zero remains centered around zero, exhibiting an increasing variance and a concomitant decrease in peak amplitude. Concurrently, the distribution of adaLN-Mix, while expanding peripherally, 
is also coalescing around zero, culminating in an unimodal structure that is symmetrically centered on zero. Though adaLN-Mix eventually overlaps with the distribution of adaLN-Zero in Fig.~\ref{fig:distribution} (e), the latter's distribution is more compact, with more values concentrated near zero. 

Essentially, adaLN-Zero exhibits a more centralized initial parameter distribution, and morphologically, its initial distribution more closely approximates the distribution observed in Fig.~\ref{fig:distribution} (e) than does the adaLN-Mix.
This could be the reason why adaLN-Zero converges faster and outperforms adaLN-Mix significantly. From an entropy perspective, our calculations show in Fig.~\ref{fig:abstract} that when adaLN-Mix is transitioning to the target distribution, \eg, from 10K steps to 50K, entropy decreases by 0.8, whereas adaLN-Zero leads to an increase in entropy. Typically, systems tend to evolve towards higher entropy (the second law of thermodynamics).  Therefore, adaLN-Zero is comparatively easier to optimize and obtains better performance.


One might question, though we have globally analyzed all $W_{\alpha}$ in DiT, is it possible that the distribution of $W_{\alpha}$ across different blocks could differ significantly from the global distribution, considering that zero-initialization is applied on a block-by-block basis? To investigate this, we examine the value distributions of $W_{\alpha}^{L}$ ($L$ is block index) of DiT-XL/2 using adaLN-Zero after training for just 10K iterations. As shown in Fig.~\ref{fig:each-block},  the distribution of $W_{\alpha}^{L}$ in each block closely resembles the pattern observed in Fig.~\ref{fig:distribution} (b), indicating that the functions of $W_{\alpha}^{L}$ across different blocks are likely analogous. This finding also supports the rationale behind \textit{uniformly} zero-initializing $W_{\alpha}^{L}$ across different blocks. 

\textbf{Remark.} Intuitively speaking, our analysis should have concluded so far. However, we observe that there are other zero-initialized modules in DiT. For the sake of completeness, we provide further analysis in the following for clarity. We also show value distributions of other non-zero-initialized DiT modules in \textcolor{red}{App.~\ref{App:More DiT Modules}} and zero convolution in ControlNet~\cite{zhang2023adding} in \textcolor{red}{App.~\ref{App:zero-convolution in ControlNet}}.

\subsection{Analysis about Other Zero-initialized Modules} \label{sec:other}

Recall that in DiT blocks, $W_{\gamma}$ and $W_{\beta}$ are zero-initialized in both adaLN-Zero and adaLN-Mix. In addition to that, the FinalLayer module is also zero-initialized at the beginning, indicating that $W_{\gamma_f}$, $W_{\beta_f}$, and $W_f$ are zero in both adaLN-Zero and adaLN-Mix. We want to investigate whether these weights exhibit behavior similar to $W_{\alpha}$.

\textbf{Analysis about $W_{\gamma}$ and $W_{\beta}$.} We present the distribution variations of the entire $W_{\gamma}$ in DiT blocks as training progresses in Fig.~\ref{fig:gamma}. It is observed that, regardless of whether it is adaLN-Zero or adaLN-Mix, $W_{\gamma}$ rapidly formulates a pattern similar to that of $W_{\alpha}$ in Fig.~\ref{fig:distribution} at a very early stage. A similar result is observed for $W_{\beta}$ as detailed in \textcolor{red}{App.~\ref{app: all beta}}. Furthermore, we also show the distribution of $W_{\gamma}^{L}$ and $W_{\beta}^{L}$ in each DiT block in \textcolor{red}{App.~\ref{app: different gamma and beta}}. Basically, the distributions of $W_{\gamma}^{L}$ and $W_{\beta}^{L}$ in each block share a similar pattern to their global ones as well as that of $W_{\alpha}$. These results indicate that $W_{\gamma}^{L}$ and $W_{\beta}^{L}$ may execute analogous functions in DiT blocks.

\textbf{Analysis about $W_{\gamma_f}$, $W_{\beta_f}$, and $W_f$.} 
As training progresses, we illustrate the variations of value distribution of $W_{\gamma_f}$ in Fig.~\ref{fig:gamma_f}, and that of $W_{\beta_f}$ and $W_f$ in \textcolor{red}{App.~\ref{app: W_f}}. We see that $W_{\gamma_f}$, $W_{\beta_f}$, and $W_f$ exhibit different tendencies. For example, $W_{\gamma_f}$ presents a bimodal distribution. These observations suggest that they may not have a consistent update direction compared to $W_{\gamma}$ and $W_{\beta}$. 

\textbf{Remark.} By comparing the results in Sec.~\ref{sec:win} and Sec.~\ref{sec:other}, we empirically demonstrate that, although the same zero-initialization strategy is used, weight distributions in different modules may also be discrepant. Meanwhile, though weights $W_{\alpha}$, $W_{\gamma}$, and $W_{\beta}$ in the conditioning mechanism are zero-initialized, after a certain number of training steps, all of them transition from zero distributions to Gaussian-like distributions~\footnote{To demonstrate that adaLN-Zero exhibits a Gaussian-like distribution, we employ KL-Divergence to measure the distance between its distribution and a true Gaussian. We use the weights of adaLN-Zero at 50K steps to compute its mean and standard deviation. These parameters are then used to initialize a Gaussian distribution, from which we sample the same number of weight points as adaLN-Zero. Therefore, the KL distance between the two sets of sampled points is calculated using the nearest neighbor non-parametric estimation method $D_{\text{KL}}(P | Q) \approx \frac{1}{n} \sum_{i=1}^{n} \log \frac{\rho_i}{\nu_i} + \log \frac{m}{n-1}$. $m$ and $n$ are the number of sample points. $\rho_i$ represents the nearest neighbor distance of point $x_{i}$ in $P$. $\nu_i$ is similar. The calculated KL-Div is 0.065. Generally, the closer the two distributions are, the smaller the KL-Div. If the two distributions are identical, the KL-Div is 0. Consequently, the computed result indicates that adaLN-Zero exhibits a Gaussian-like distribution. This similarity may be influenced by the denoising task, which gradually removes Gaussian noise. As our focus is not on the reasons behind these patterns, we leave this exploration as future work.}. This characteristic inspires us to directly initialize these weights with a suitable Gaussian distribution to accelerate training, which we put in the next section to verify.

\section{adaLN-Gaussian}

Our insight is that, as training progresses, the weight distribution gradually transitions from zero to a Gaussian-like distribution. Thus, why do we not directly initialize the weights with a Gaussian distribution to potentially expedite this distribution shift and accelerate training?

\setlength{\tabcolsep}{0.8cm}{\begin{table}[h]
\centering
\caption{Results of different std settings. $0$: adaLN-Zero}
\begin{tabular}{l c c}
	\toprule
        Std & FID  & IS
   \\
	\midrule
 0 &  78.99 & 14.19  \\
\midrule
 5e-4 & 80.68 & 13.93  \\
 8e-4 & 79.49 & 14.54 \\
 \rowcolor{yellow!30} 1e-3 & \textbf{76.21} & \textbf{15.01} \\
 2e-3 & 78.91  & 14.33 \\
 5e-3 & 79.54 & 14.33 \\
 5e-2 & 84.37 & 13.67 \\
\bottomrule
\end{tabular}
\label{tab:corse_std}
\vskip -0.1in
\end{table}}

\setlength{\tabcolsep}{0.5cm}{\begin{table}
\centering
\caption{Ablation study for $W_{\alpha}$. (0, 0, 0): adaLN-Zero}
\begin{tabular}{l c c}
	\toprule
        Std ($W_{\alpha}$, $W_{\gamma}$, $W_{\beta}$) & FID  & IS
   \\
	\midrule
 0, 0, 0 & 78.99 & 14.19  \\
 1e-3, 0, 0 & 78.62 & 14.42 \\
 1e-3, 1e-3, 1e-3 & 76.21 & 15.01 \\
\bottomrule
\end{tabular}
\label{tab:ablation}
\vskip -0.1in
\end{table}}

To leverage Gaussian distribution to initialize $W_{\alpha}$, $W_{\gamma}$, and $W_{\beta}$, we need to determine the appropriate standard deviation (std), with the mean value defaulting to 0. Intuitively, we can determine the std value by approximating the weight distribution at a specific moment during the training of adaLN-Zero. Moreover, this moment should be neither too late, as initializing $W_{\alpha}$, $W_{\gamma}$, and $W_{\beta}$ at a later stage may impart learned priors incompatible with vanilla weights, nor too early, as there may be minimal difference from zero-initialization (In \textcolor{red}{App.~\ref{App:Analysis-Different-Std}}, we give a detailed result analysis about different std choices in Gaussian initialization.). Therefore, based on Fig.~\ref{fig:distribution}, we heuristically select and ablate several std values to uniformly initialize $W_{\alpha}$, $W_{\gamma}$, and $W_{\beta}$ and train each variant for 50K iterations for simplicity. The results are presented in Tab.~\ref{tab:corse_std} where $std=1e-3$ yields the best performance among all variants. This result verifies the effectiveness of our idea and suggests that Gaussian initialization with appropriate parameters is able to outperform zero-initialization under the same steps (\ie, converging faster). We denote this initialization method as \textit{adaLN-Gaussian}. The pytorch implementation below is simple with only one line replaced.

\lstset{
    language=Python,               
    basicstyle=\ttfamily,          
    keywordstyle=\color{blue},     
    commentstyle=\color{green},   
    stringstyle=\color{red},       
    numbers=left,                  
    numberstyle=\tiny\color{gray}, 
    frame=single,                  
    breaklines=true,               
}
\begin{scriptsize}
\begin{lstlisting}[language=Python, escapeinside={(*@}{@*)}]
for ind, block in enumerate(self.blocks):
    nn.init.constant_(block.adaLN_modulation[-1].bias, 0)
    (*@\st{$\operatorname{nn.init.constant\_(block.adaLN\_modulation[-1].weight, 0)}$}@*)
    nn.init.normal_(block.adaLN_modulation[-1].weight, std=0.001)
\end{lstlisting}
\end{scriptsize}

\setlength{\tabcolsep}{0.45cm}{\begin{table*}[htbp]
\vskip -0.05in
\centering
        \caption{Comparison on longer training time, different DiT variants, larger image size, and more DiT-based models. We additionally report sFID~\cite{nash2021generating} and Precision/Recall~\cite{kynkaanniemi2019improved}
as secondary metrics following DiT. CFG: Classifier-free guidance. For CFG, we use DiT-XL/2's best guidance value. We use ImageNet1K $256\times256$ by default if not specified. $^*$ indicates that the results are borrowed from ~\cite{mo2023dit}}
			\begin{tabular}{l l c l c c c c c}
				\toprule
				Model & Initialization & CFG & Steps &  FID$\downarrow$  & sFID$\downarrow$ & IS$\uparrow$ & Precision$\uparrow$ & Recall$\uparrow$ \\ 
				\midrule
                    \midrule
                    \multicolumn{4}{l}{\emph{Longer training time}:}\\
				DiT-XL/2 &  adaLN-Zero  & 1 & 400K  & 20.02  & 6.09 & 67.34 & 63.33 & 63.06  \\
    			 \rowcolor{yellow!20} DiT-XL/2
				& adaLN-Gaussian & 1 & 400K  & \textbf{17.86}  &  6.06 & 73.07 & 64.51 & 62.64  \\
                    DiT-XL/2  
				&  adaLN-Zero & 1 & 800K  & 14.73  & 6.35 & 86.70 & 65.62 & 63.93 \\
    			 \rowcolor{yellow!20} DiT-XL/2
				& adaLN-Gaussian & 1 & 800K  & \textbf{13.14}  & 6.11 & 92.98 & 66.50 & 63.92 \\
                                    DiT-XL/2$^*$ 
				&  adaLN-Zero & 1 & 2352K  &   10.67 & - & - & - & - \\
    			 \rowcolor{yellow!20} DiT-XL/2
				& adaLN-Gaussian & 1 & 2350K  &   \textbf{10.28} & 6.48 & 112.76 & 67.31 & 65.39 \\
                DiT-XL/2  
				&  adaLN-Zero & 1.5 & 400K  &  6.15 & 4.60  & 152.70 & 79.92 & 52.28  \\
    			 \rowcolor{yellow!20} DiT-XL/2
				& adaLN-Gaussian & 1.5 & 400K  & \textbf{5.28}  & 4.62 & 164.62 & 80.75 & 52.65 \\
			 DiT-XL/2$^*$ 	&  adaLN-Zero & 1.5 & 7000K  & 2.27 & 4.60 & 278.24 & 83.00 & 57.00 \\
             \rowcolor{yellow!20} DiT-XL/2
				& adaLN-Gaussian & 1.5 & 3800K  & 2.27 & 5.02 & 275.63 & 79.61 & 60.60  \\ 
    			 \rowcolor{yellow!20} DiT-XL/2
				& adaLN-Gaussian & 1.5 & 4400K  & 2.26 & 4.94 & 276.11 & 80.35 & 60.70  \\ 
           \rowcolor{yellow!20} DiT-XL/2 & adaLN-Gaussian & 1.5 & 5000K & 2.23 & 4.90 & 277.55 & 80.58 &  59.90 \\
           \rowcolor{yellow!20} DiT-XL/2 & adaLN-Gaussian & 1.5 & 6000K  & 2.22 & 4.94 &  275.93 & 80.35 & 60.24\\
            \rowcolor{yellow!20} DiT-XL/2 & adaLN-Gaussian & 1.5 & 7000K  & \textbf{2.21} & 4.92 &  275.89 & 79.97 & 60.85 \\
				\midrule 
    \multicolumn{4}{l}{\emph{Different DiT variants and larger image size}:} \\
        DiT-B/2
				& adaLN-Zero & 1 & 400K & 42.72 & 8.29 & 33.28 & 
 49.02 & 62.80 \\
     \rowcolor{yellow!20} DiT-B/2
				& adaLN-Gaussian & 1 & 400K & \textbf{42.55} & 8.13 & 33.82 & 49.05 & 63.30 \\
				DiT-L/2
				& adaLN-Zero & 1 & 400K & 24.40 & 6.47 & 57.47 & 60.14 & 63.21 \\
     \rowcolor{yellow!20} DiT-L/2
				& adaLN-Gaussian & 1 & 400K & \textbf{23.05} & 6.39 & 60.49 & 61.44 & 62.27 \\
                    DiT-L/4
				& adaLN-Zero & 1 & 400K & 45.71 & 9.26 & 32.00 & 46.61 & 60.71 \\
     \rowcolor{yellow!20} DiT-L/4
				& adaLN-Gaussian & 1 & 400K & \textbf{44.11} & 9.06 & 33.13 & 47.51 & 61.42 \\
    
    DiT-XL/4$_{512\times512}$ &  adaLN-Zero & 1 & 400K  &  35.21 & 8.00 & 42.42 & 65.87 & 62.70  \\
    			 \rowcolor{yellow!20} DiT-XL/4$_{512\times512}$
				& adaLN-Gaussian & 1 & 400K  & \textbf{34.68}  & 7.86 & 42.75 & 65.95 & 61.90  \\
                \midrule
    \multicolumn{4}{l}{\emph{Compatibility with other improving methods}:} \\
        FasterDiT-XL/2 & adaLN-Zero & 1 & 400K & 12.64 & 5.13 & 93.15 & 65.94 & 64.56 \\
     \rowcolor{yellow!20} FasterDiT-XL/2 & adaLN-Gaussian & 1 & 400K & \textbf{12.45} & 5.10 & 94.42 & 65.62 & 64.98 \\
				SiT-XL/2
				& adaLN-Zero & 1 & 400K & 18.97 & 5.23 & 71.06  & 62.82 & 63.49 \\
     \rowcolor{yellow!20} SiT-XL/2
				& adaLN-Gaussian & 1 & 400K & \textbf{18.66} & 5.20 & 71.60 & 62.61 & 64.05 \\
				\midrule
    \multicolumn{4}{l}{\emph{Different DiT-based models}:} \\
				LlamaVision-XL/2
				& adaLN-Zero & 1 & 400K & 21.66 & 6.61 & 65.66 & 60.78 & 63.78 \\
     \rowcolor{yellow!20} LlamaVision-XL/2
				& adaLN-Gaussian & 1 & 400K & \textbf{20.26} & 6.20 & 68.82 & 62.06 & 63.90 \\
    U-DiT-L & adaLN-Zero & 1 & 400K & 10.87 & 5.35 & 108.08 & 70.64 & 61.74 \\
     \rowcolor{yellow!20} U-DiT-L & adaLN-Gaussian & 1 & 400K & \textbf{10.47} & 5.37 &  108.58 & 70.68 & 61.82 \\
				\bottomrule
			\end{tabular}
            \vskip -0.15in
		\label{tab:comprehensive comparison}
\end{table*}}

Additionally, we conduct an ablation in Tab.~\ref{tab:ablation} where we apply Gaussian initialization only for $W_{\alpha}$~\footnote{We observe that $W_{\alpha}$ plays a critical role in adaLN-Zero compared to adaLN, with its initial value significantly impacting model performance (adaLN-Zero \vs. adaLN-Step1). Thus, we primarily ablate $W_{\alpha}$ rather than $W_{\gamma}$ and $W_{\beta}$.}. This is the same as adaLN-Step1 but adaLN-Step1 uses default initialization for $W_{\alpha}$. Hence we denote this variant as adaLN-Step1-Gaussian. Recall that adaLN-Step1 is remarkably inferior to adaLN-Zero while adaLN-Step1-Gaussian here unexpectedly matches and even outperforms adaLN-Zero. \textit{This supports our hypothesis that a good initialized position in the optimization space is the key.} It also indicates that zero initialization may not be the best choice.


Though the distributions of $W_{\alpha}$, $W_{\gamma}$, and $W_{\beta}$ all resemble Gaussian distribution, in Fig.~\ref{fig:distribution} (b), Fig.~\ref{fig:gamma} (b), and \textcolor{red}{App.~\ref{app: all beta}} Fig.~\textcolor{red}{3} (b)  discrepancies in their shapes persist, \eg, bottom width. Thus, it is more appropriate to select std for each of them independently. We perform a grid search and empirically find that $std(8e-4, 1.2e-3, 8e-4)$ produces the best FID. We denote this initialization as \textit{adaLN-Gaussian-v2} and include the search results of adaLN-Gaussian-v2 in \textcolor{red}{App.~\ref{app:adaLN-Gaussian-v2}} for clarity.

\textbf{Longer training time.}  To verify the effectiveness of our initialization strategies, as shown in Tab.~\ref{tab:comprehensive comparison}, we train DiT-XL/2 with long training steps including 400K and 800K on ImageNet1K $256\times256$ w/wo CFG. One can see that adaLN-Gaussian outperforms adaLN-Zero under the same steps by a large margin, demonstrating the efficiency of our initialization strategies. We show more results in Fig.~\ref{fig:fid}. We further extend the training steps to 2350K and 7000K following DiT to evaluate performance w/wo CFG. It is seen that adaLN-Gaussian still outperforms adaLN-Zero in 2350K.  As training progresses, the impact of initialization gradually diminishes, making it reasonable for the performance gap to narrow. On the other hand, we observe that our method basically converges at 3800K training steps and matches the converged performance of adaLN-Zero (2.27 FID), providing a 46\% time savings. After further training, our method outperforms adaLN-Zero and yields 2.21 FID at 7000K~\footnote{It is worth noting that adaLN-Gaussian does not alter the model architecture or the learning algorithm of DiT model, which means that the model's capacity is the same. Therefore, theoretically, given enough training time, adaLN-Gaussian could not bring very \textit{significant} improvements on the final performance, but converge faster.}. These results collectively demonstrate the advantage of our initialization strategies on improving training efficiency. We also show the results of adaLN-Gaussian-v2 in \textcolor{red}{App.~\ref{app:adaLN-Gaussian-v2}}. 

\textbf{Generalization to different DiT variants and larger image size.} To demonstrate the adaLN-Gaussian is a general method, we conduct experiments on several commonly-used DiT variants including  DiT-B/2, DiT-L/2, and DiT-L/4. As shown in Tab.~\ref{tab:comprehensive comparison}, we see that adaLN-Gaussian also improves the performance of DiT-B/2, DiT-L/2, and DiT-L/4 though its parameter is set according to DiT-XL/2 and may not be the best setting for these three variants. We further demonstrate the generalization on ImageNet1K $512\times512$. These results show the effectiveness of adaLN-Gaussian and imply the great potential of our method after more precise case-by-case adjustments.

\textbf{Compatibility with other improving methods.} Additionally, we recognize that there are several studies that focus on improving DiT training, such as improvements in training framework and algorithm. To assess the compatibility of adaLN-Gaussian with these methods, we select one representative method from each category: SiT~\cite{ma2024sit} and FasterDiT~\cite{yaofasterdit}. As shown in Tab~\ref{tab:comprehensive comparison}, we employ adaLN-Gaussian on these methods and the results show the superiority of adaLN-Gaussian over adaLN-Zero, demonstrating the compatibility of adaLN-Gaussian.

\textbf{Generalization to other DiT-based models and datasets~\footnote{To save GPU memory, we use the fast version of DiT Github code (\url{https://github.com/chuanyangjin/fast-DiT}) featuring gradient checkpointing, mixed precision training, and pre-extracted VAE features, all of which are employed in experiments of Tab.~\ref{tab:comprehensive comparison} if not specified. Consequently, though we follow all the training settings, the reported results may be slightly different from that of the original paper.}.} We further validate the effectiveness of our method across different enhanced DiT-based models, including those incorporating advanced architectures and integrating specific priors. To this end, we select LlamaVision~\cite{chu2024visionllama} and U-DiT~\cite{tian2024u} as representative models, corresponding to these two aspects, respectively. As presented in Tab.~\ref{tab:comprehensive comparison}, adaLN-Gaussian exhibits higher training efficiency than adaLN-Zero, demonstrating the generalization of adaLN-Gaussian. Moreover, we also show the effectiveness and generalization of adaLN-Gaussian compared to adaLN-Zero on more datasets including Tinyimagenet~\cite{le2015tiny}, AFHQ~\cite{choi2020stargan}, and CelebA-HQ~\cite{karras2018progressive} in \textcolor{red}{App.~\ref{App:more-experiments}}.

\textbf{Effectiveness on text-to-image generation.} Beyond class-to-image generation, we also evaluate the effectiveness of our method on text-to-image generation task. Specifically, built on DiT-XL/2, we leverage CLIP text encoder~\cite{radford2021learning} for text encoding and insert cross-attention to each transformer block (between self-attention and FFN module) to incorporate text conditioning. We use 8 H800 GPUs to train the model on LAION-Aesthetics dataset (score over 6.25) for 50K steps for simplicity (around 2 epochs). We set batch size to 128 while maintaining other settings. The performance is evaluated on COCO FID-30K and CLIP score following previous efforts~\cite{rombach2022high, esser2024scaling}. The results are shown in Tab~\ref{tab:text2image}. One can see that under the same 50K steps, adaLN-Gaussian achieves 65.51 FID, outperforming adaLN-Zero (71.41 FID) by a large margin. Moreover, our method obtains 0.2178 CLIP score and also outperforms adaLN-Zero. These experiments together demonstrate the effectiveness and generalization of our method on improving training efficiency for text-to-image generation.

\setlength{\tabcolsep}{0.4cm}{\begin{table}[h]
		\caption{Performance comparison on text-to-image generation task.} 
        \vskip -0.2in
		\begin{center}
        \small
			\begin{tabular}{l c c c c c c c}
				\toprule
				   Method &  FID30K$\downarrow$ & CLIP score$\uparrow$ & \\ 
				\midrule
				adaLN-Zero & 71.41 & 0.2143 \\
                adaLN-Gaussian & 65.51 & 0.2178  \\
				\bottomrule
			\end{tabular}
		\end{center}
            \vskip -0.1in
		\label{tab:text2image}
\end{table}}

\textbf{Discussion.} Although the improvement of adaLN-Gaussian on the final converged performance is relatively modest after extremely long training, this observation is expected since adaLN-Gaussian is ensensially an initialization strategy without changing model capacity. This observation is also consistent with the general understanding that initialization primarily affects optimization efficiency rather than the ultimate representational capacity of sufficiently trained models. The primary contribution of adaLN-Gaussian is to provides a better optimization starting point    and improve training efficiency by helping model converge faster.  This can lead to practical benefits like saving computational cost, which is valuable in time and resource-limited scenarios, especially in the era of large model. On the other hand,  the primary significance of adaLN-Gaussian is not the absolute improvement of the final performance bound, but that it provides an analysis-driven validation of the findings uncovered in this work: The experimental results of adaLN-Gaussian indicate that compared to adaLN-Zero, adaLN-Gaussian provides a more suitable initialization that adapts to the evolving weight distribution more quickly. To some extent, our analysis and the resulting initialization strategy demonstrate how understanding adaLN-Zero can naturally lead to improved initialization strategies. Furthermore, the insights obtained from this analysis are not limited to adaLN-Gaussian. They also inspire the design of the SE-adaLN-Zero introduced in the following section.

\setlength{\tabcolsep}{0.4cm}{\begin{table}[!htbp]
\centering
\caption{Comparison of different SE-like variants with different ratios compared to adaLN-Zero using DiT-XL/2.}
\begin{tabular}{l | c c c c}
\toprule
Method & Ratio & Params &  FID $\downarrow$ & IS$\uparrow$ \\
\midrule
adaLN-Zero & 1 & 676M & 78.99  & 14.19  \\
 \midrule
SE-like v1 & 2 & 562M & 77.09  & 15.20 \\
SE-like v1 & 4 & 506M & 82.01 & 13.97 \\
SE-like v1 & 8 & 477M & 84.35 & 13.17 \\
 \midrule
SE-like v2 & 2 & 582M & \textbf{76.50} & \textbf{15.96} \\
SE-like v2 & 4 & 517M & 79.06 & 14.61 \\
SE-like v2 & 8 & 484M & 82.65 & 13.69 \\
\midrule
SE-like v3 & 2 & 560M & 89.82 & 12.67 \\
SE-like v3 & 4 & 503M & 91.76 & 12.10 \\
\bottomrule
\end{tabular}
\label{tab:SE-variant}
\end{table}

\begin{figure}[htbp]
\centering{\includegraphics[width=1.0\linewidth]{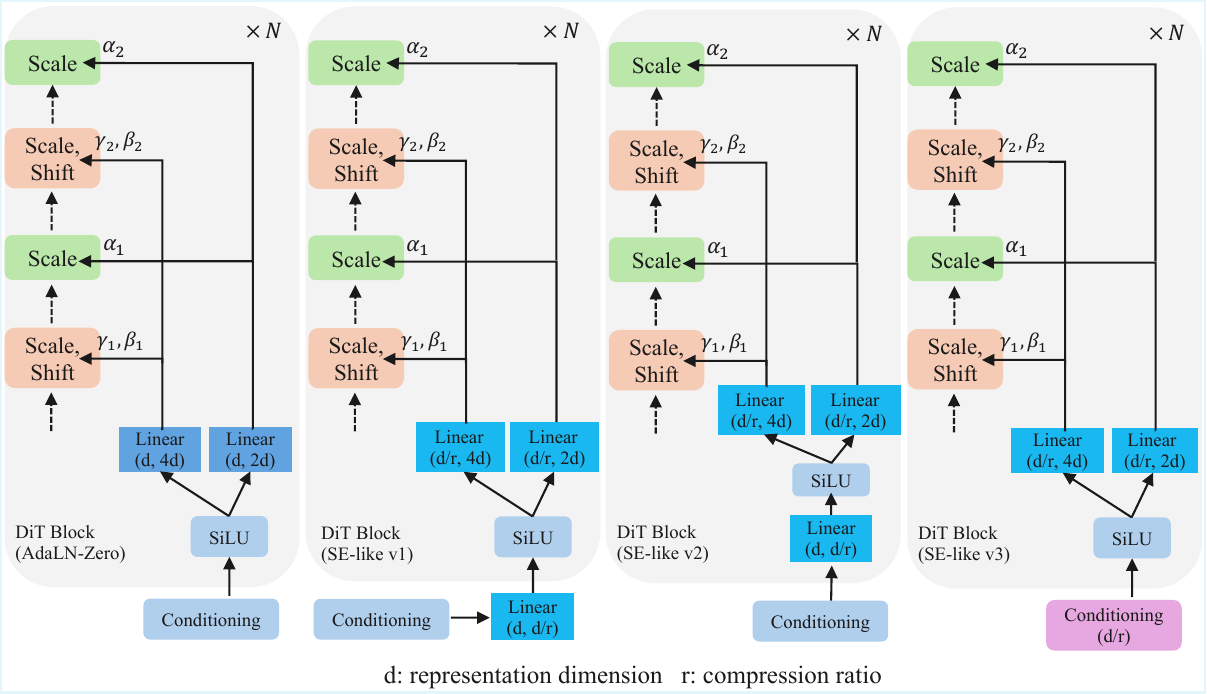}}
	\caption{Detailed illustration of our SE-like structure variants.}
	\label{fig:SE-like-structure}
\end{figure}

\setlength{\tabcolsep}{0.35cm}{\begin{table*}[!htbp]
\centering
\caption{Comparisons between adaLN-Zero and our SE-adaLN-Zero variant under longer training time, different DiT size, and more DiT-based models. We
report FID, sFID, IS, and Precision/Recall following DiT. CFG: Classifier-free guidance.}
\begin{tabular}{l | c c c c c c c c c}
\toprule
Model & Type & Params & CFG &Steps & FID $\downarrow$ & sFID$\downarrow$  & IS$\uparrow$ & Precision$\uparrow$ & Recall$\uparrow$ \\
\midrule
DiT-XL/2 & adaLN-Zero & 676M & 1 &400K & 20.02 & 6.09 & 67.34 & 63.33 & 63.06  \\
\rowcolor{yellow!20} DiT-XL/2 & SE-adaLN-Zero & 582M & 1 & 400K & \textbf{19.13} & 6.08 & 68.85 & 63.64 & 63.59 \\
\rowcolor{yellow!20} DiT-XL/2 & SE-adaLN-Gaussian & 582M & 1 & 400K & \textbf{18.76} & 5.85 & 68.62 & 64.40 & 62.38 \\
DiT-XL/2 & adaLN-Zero & 676M & 1.5 &400K & 6.15 & 4.60 & 152.70 & 79.92 & 52.28  \\
\rowcolor{yellow!20} DiT-XL/2 & SE-adaLN-Zero & 582M & 1.5 & 400K & \textbf{5.67} & 4.62 & 154.61 & 79.80 & 53.31 \\
DiT-XL/2 & adaLN-Zero & 676M & 1 & 800K & 14.73 & 6.35 & 86.70 & 65.62 & 63.93 \\
\rowcolor{yellow!20} DiT-XL/2 & SE-adaLN-Zero & 582M & 1 & 800K & \textbf{13.52} & 6.08 & 88.59 & 66.45 & 64.52 \\
DiT-XL/2 & adaLN-Zero & 676M & 1 & 2352K & 10.67  &  - & - & - & - \\
\rowcolor{yellow!20} DiT-XL/2 & SE-adaLN-Zero & 582M & 1 & 2350K & \textbf{10.26} & 6.18 & 109.41 & 66.92 & 66.52 \\
DiT-XL/2 & adaLN-Zero & 676M & 1.5 & 7000K &   2.27 & 4.60 & 278.24 & 83.00 & 57.00 \\
\rowcolor{yellow!20} DiT-XL/2 & SE-adaLN-Zero & 582M & 1.5 & 4500K & \textbf{2.26} & 4.89 & 272.92 & 79.88 & 60.13 \\
\rowcolor{yellow!20} DiT-XL/2 & SE-adaLN-Zero & 582M & 1.5 & 7000K & \textbf{2.21} & 4.84 & 270.15 & 80.21 & 59.93 \\
\midrule
DiT-L/2 & adaLN-Zero & 458M & 1 & 400K & 24.40 & 6.47 & 57.47 & 60.14 & 63.21 \\
\rowcolor{yellow!20} DiT-L/2 & SE-adaLN-Zero & 395M & 1 & 400K & \textbf{22.12} & 6.24 & 62.15 & 61.73 & 63.16 \\
DiT-B/2 & adaLN-Zero & 130M & 1 & 400K & 42.72 & 8.29 & 33.28 & 49.02 & 62.80 \\
\rowcolor{yellow!20} DiT-B/2 & SE-adaLN-Zero & 112M & 1 & 400K & \textbf{42.47} & 7.95 & 33.15 & 49.06 & 62.14 \\
\midrule
FasterDiT-XL/2 & adaLN-Zero & 676M & 1 & 400K & 12.64 & 5.13 & 93.15 & 65.94 & 64.56\\
\rowcolor{yellow!20} FasterDiT-XL/2 & SE-adaLN-Zero & 582M & 1 & 400K & \textbf{12.46} & 5.07 & 93.65 & 65.93 & 64.79 \\
\midrule
LlamaVision-XL/2 & adaLN-Zero & 676M & 1 & 400K & 21.66 & 6.61 & 65.66 & 60.78 & 63.78 \\
\rowcolor{yellow!20} LlamaVision-XL/2 & SE-adaLN-Zero & 582M & 1 & 400K & \textbf{18.61} & 5.90 & 71.74 & 63.32 & 63.44 \\
\bottomrule
\end{tabular}
\vskip -0.05in
\label{tab:SE-evaluate}
\end{table*}

\section{SE-adaLN-Zero}

Besides adaLN-Gaussian, we also introduce another improved conditioning mechanism. Specifically, inspired by our analysis of the SE-like structure in Sec.~\ref{sec:SE_analysis}, we consider two structure variants (SE-like v1 and v2) for adaLN-Zero that more closely resemble the SE architecture~\cite{hu2018squeeze}. Additionally, motivated by PixArt-$\alpha$~\cite{chen2023pixart}, we further consider another variant (SE-like v3). We present their structures as well as adaLN-Zero for better comparison as illustrated in Fig~\ref{fig:SE-like-structure}.

In Tab~\ref{tab:SE-variant}, we evaluate these SE-like variants with different ratios compared to adaLN-Zero under 50K steps training using ImageNet1K 256$\times$256 based on the largest model DiT-XL/2. It can be seen that SE-like variants can significantly reduce model parameters because the adaLN-Zero module of the DiT-XL/2 accounts for a substantial proportion (66\%) of the parameters. On the other hand, we find that a suitable SE-like structure and ratio can maintain and even outperform the baseline model. For example, our SE-like v1 ($r=2$) produces 77.09 for FID compared to adaLN-Zero (78.99 FID). Our SE-like v2 uses $r=4$ to reduce more parameters (around 24\%) while keeping similar performance on FID compared to adaLN-Zero. Additionally, we notice that SE-like v2 ($r=2$) obtains the best performance among all models and also reduces around 14\% parameters. Hence, we adopt this variant and call it as \textit{SE-adaLN-Zero} for simplicity.

Similar to Tab~\ref{tab:comprehensive comparison}, we comprehensively compare our SE-adaLN-Zero with adaLN-Zero on ImageNet1K 256$\times$256 in Tab~\ref{tab:SE-evaluate}. We train with longer training steps from 400K to 7000K w/wo CFG. SE-adaLN-Zero is still  superior over adaLN-Zero. For instance, SE-adaLN-Zero lowers the FID to 19.13 (\vs 20.02 for adaLN-Zero) at 400K, and further improves to 2.21 at 7000K (\vs 2.27). We additionally incorporate our Gaussian initialization into SE-adaLN-Zero to demonstrate their compatibility. This combined method, which we term SE-adaLN-Gaussian, further improves upon SE-adaLN-Zero by lowering the FID from 19.13 to 18.76. Also, we demonstrate the generalization of SE-adaLN-Zero to different model sizes such as DiT-L/2 and DiT-B/2. Besides, we further show the compatibility and generalization to other improving methods (FasterDiT) and DiT-based models (LlamaVision).

\setlength{\tabcolsep}{0.3cm}{\begin{table}[!htbp]
		\caption{Performance comparison on text-to-image generation task.} 
		\begin{center}
        \small
			\begin{tabular}{l c c c c c c c}
				\toprule
				   Method & Params & FID30K$\downarrow$ & CLIP score$\uparrow$ & \\ 
				\midrule
				adaLN-Zero & 826M & 71.41 & 0.2143 \\
                SE-adaLN-Zero & 733M & 69.18 & 0.2156 \\
				\bottomrule
			\end{tabular}
		\end{center}
		\label{tab:SE-text2image}
\end{table}}

Finally, we also evaluate SE-adaLN-Zero on text-to-image generation task to further show its generalization. Specifically, by following the same settings in Tab~\ref{tab:text2image}, we report the COCO FID-30K and CLIP score in Tab~\ref{tab:SE-text2image}. One can see that our SE-adaLN-Zero achieves 69.18 FID and 0.2156 CLIP score, outperforming adaLN-Zero in both metrics while using less model parameters, verifying the effectiveness of SE-adaLN-Zero.

\section{Conclusion}
We study three key factors contributing to the performance discrepancy: an SE-like structure, a good zero-initialized value, and a "gradual" update order of model weights. Moreover, our empirical experiments suggest that a good zero-initialized value \textit{itself} plays a more significant role among these factors. Based on the observed distribution variations in condition modulation weights, we propose adaLN-Gaussian which uses Gaussian distributions to initialize condition modulations. On the other hand, inspired by the analysis of SE-like structure, we additionally propose an enhanced conditioning mechanism called SE-adaLN-Zero. We conduct extensive experiments with DiT on four datasets, especially on ImageNet1K across different settings and text-to-image experiments, demonstrating the effectiveness and generalization of adaLN-Gaussian and SE-adaLN-Zero. Importantly, beyond the numerical gains brought by the proposed methods, we believe that our work provides a clearer understanding of why adaLN-Zero works and illustrates how such understanding can guide the design of improved conditioning mechanism.


\bibliographystyle{IEEEtran}
\bibliography{bibfile}

\begin{IEEEbiography}[{\includegraphics[width=1in,height=1.25in,clip,keepaspectratio]{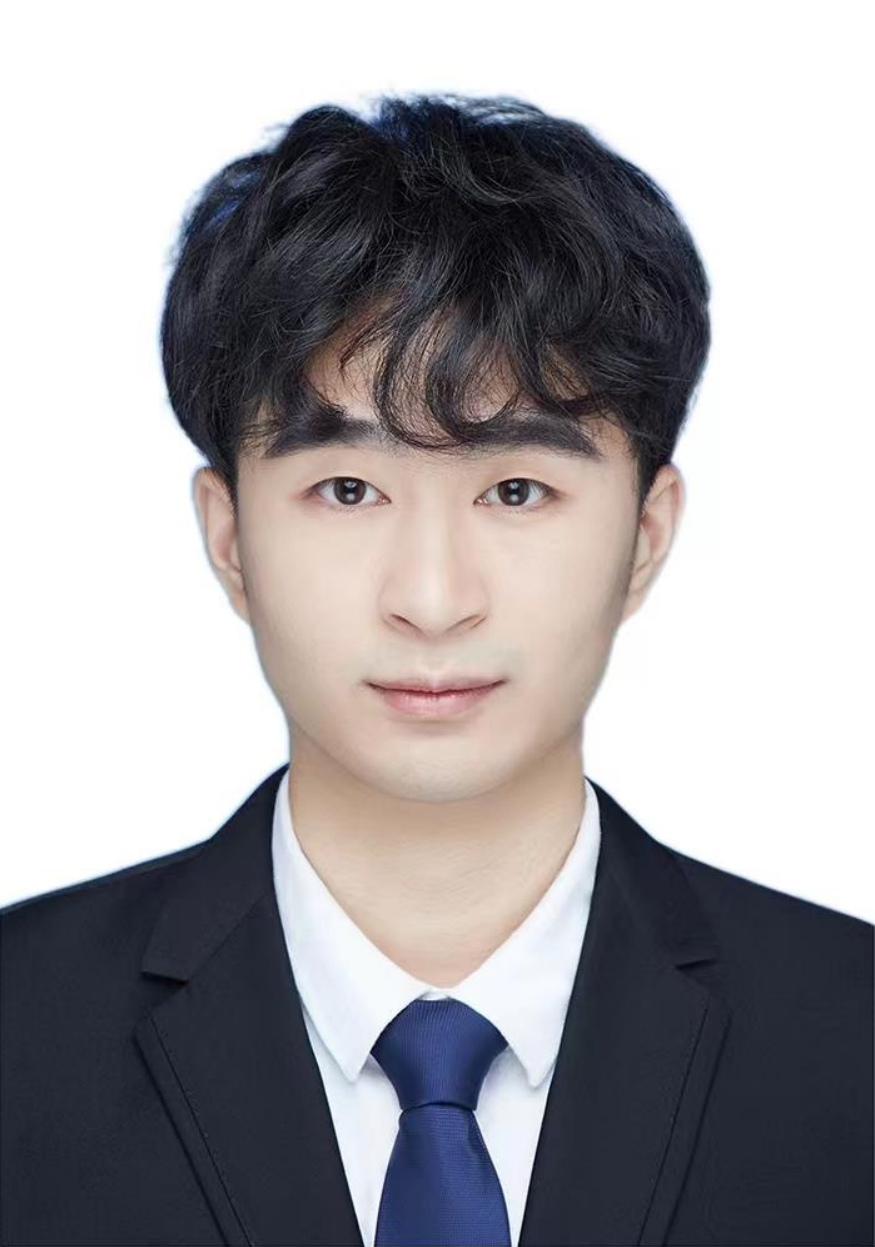}}]{Jie Zhu}
is currently working toward the Ph.D. degree in the School of Computer Science, Peking University, China.
\end{IEEEbiography}

\begin{IEEEbiography}[{\includegraphics[width=1in,height=1.25in,clip,keepaspectratio]{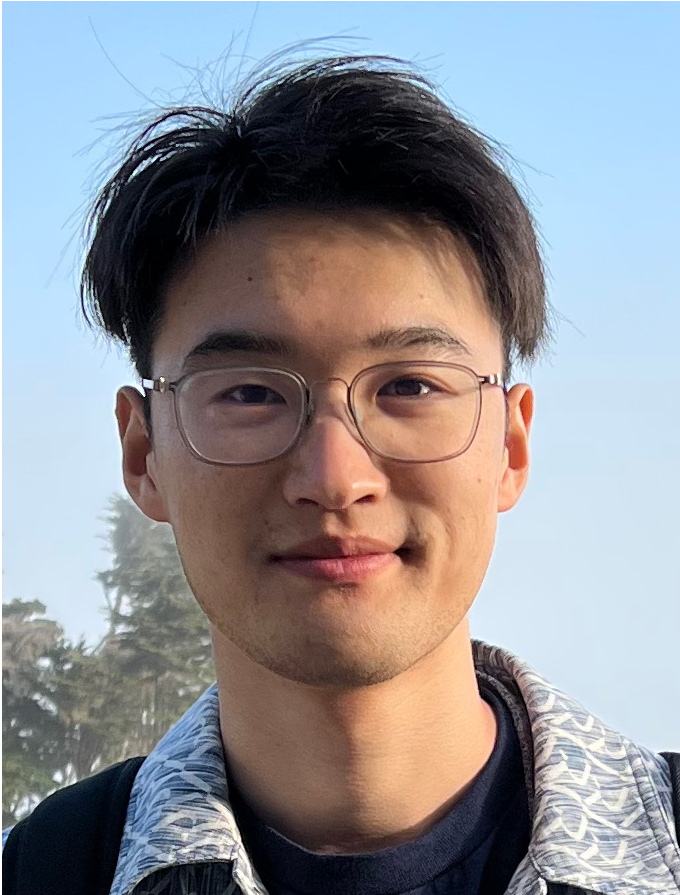}}]{Mingyu Ding} is an assistant professor of computer science at UNC-Chapel Hill. He was a postdoctoral fellow at BAIR, UC Berkeley and obtained a doctorate from the University of Hong Kong in 2022. Mingyu's research interests lie at the intersection of robotics, embodied AI, and computer vision. His works have been recognized by several research fellowships and 2 best paper awards at ICRA.
\end{IEEEbiography}

\begin{IEEEbiography}[{\includegraphics[width=1in,height=1.25in,clip,keepaspectratio]{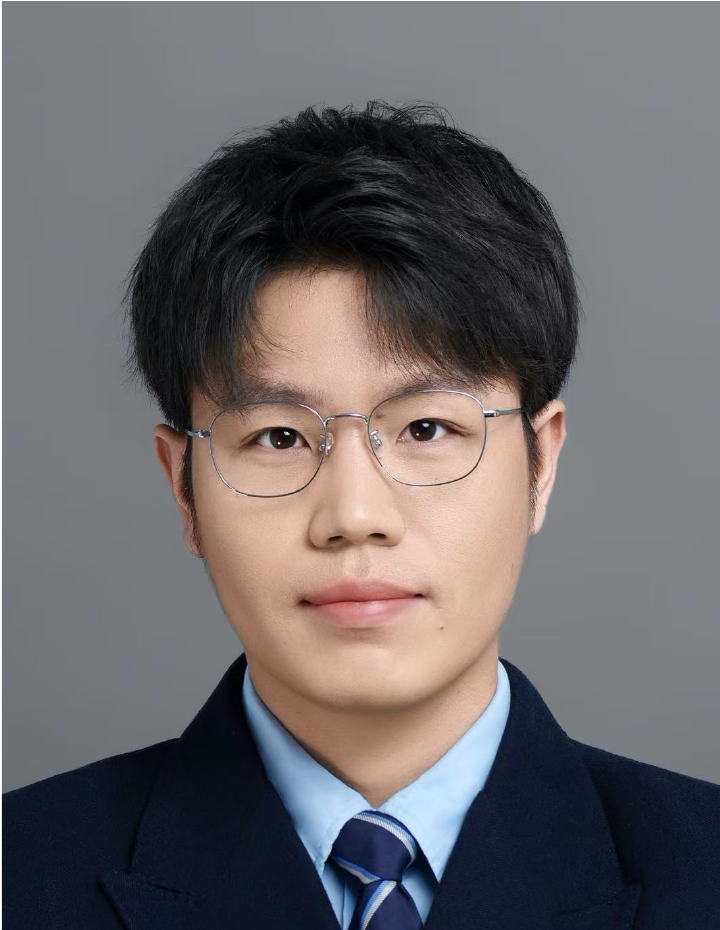}}]{Boqiang Duan} is currently a Senior Research Engineer in Department of Computer Vision Technology (VIS), Baidu Inc., Beijing, China. His current research interests include computer vision and generative AI.
\end{IEEEbiography}

\begin{IEEEbiography}[{\includegraphics[width=1in,height=1.25in,clip,keepaspectratio]{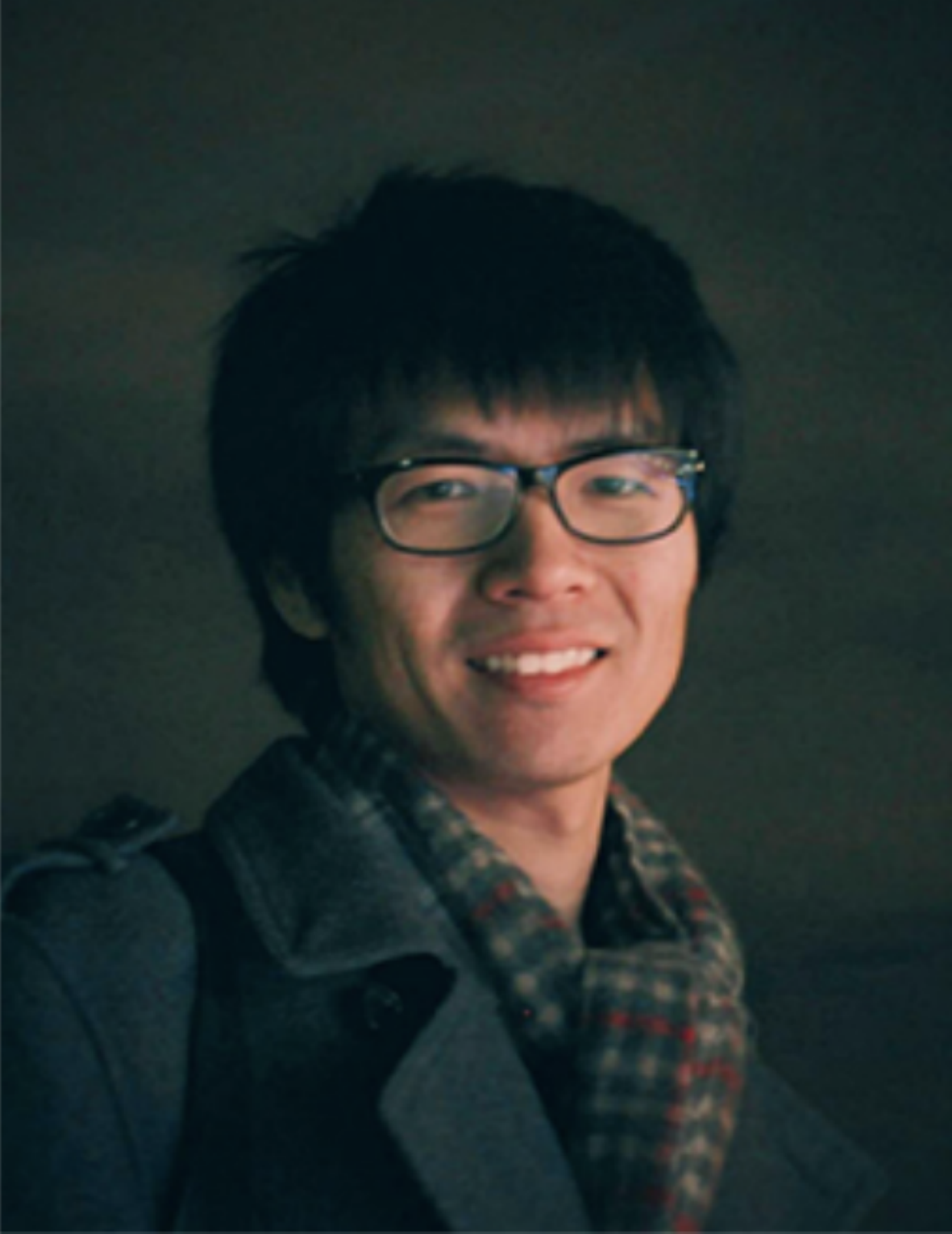}}]{Leye Wang} 
received the Ph.D. degree in computer science from TELECOM SudParis and University Paris 6, France, in 2016. He is currently a tenured associate professor with the Key Lab of High Confidence Software Technologies, Peking University, MOE, and the School of Computer Science,
Peking University, China. He was a postdoctoral
researcher with the Hong Kong University of Science and Technology. His research interests
include ubiquitous computing, mobile crowdsensing, and urban computing.
\end{IEEEbiography}

\begin{IEEEbiography}[{\includegraphics[width=1in,height=1.25in,clip,keepaspectratio]{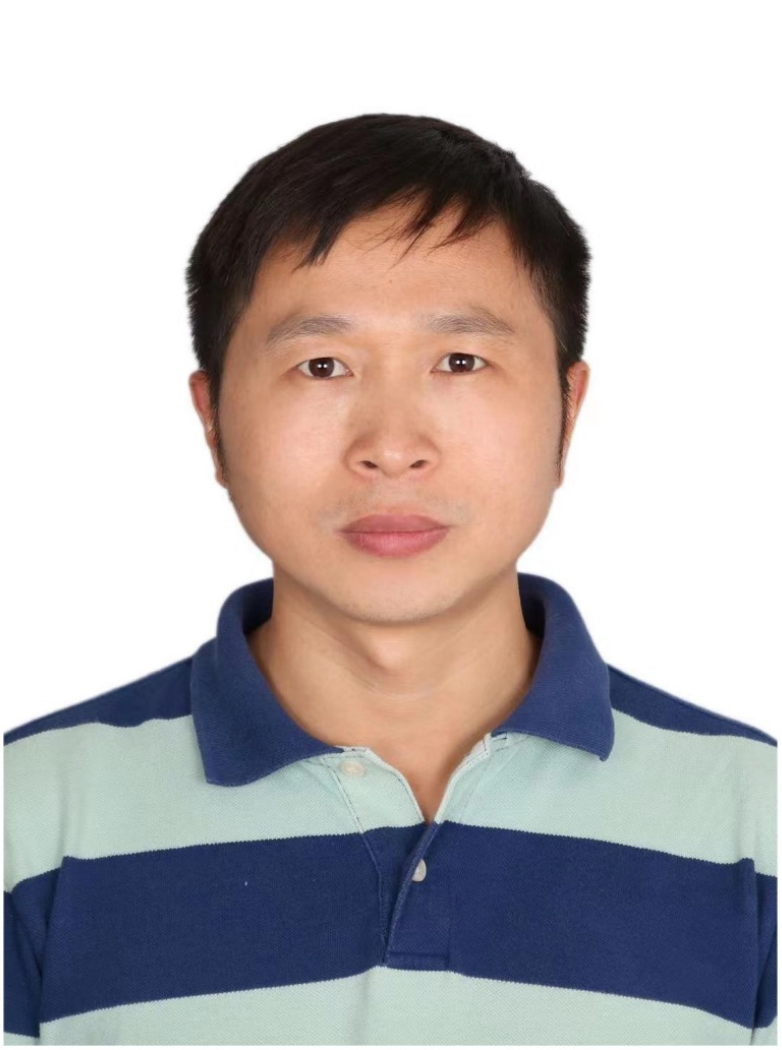}}]{Jingdong Wang}  is a Fellow of CAE, IEEE, IAPR, and AAIA, and an ACM Distinguished Member. He was Chief Scientist for Computer Vision at Baidu and a Senior Principal Researcher at Microsoft Research Asia. His current research interests lie in multimodal understanding and generation, diffusion models, and world models. His work spans neural architecture design, semantic segmentation, and large‑scale vector similarity search. He served as Program Chair of ICCV 2025. He frequently serves as a (Senior) Area Chair for major conferences including CVPR, ICCV, ECCV, NeurIPS, ICML, ICLR, ACM MM, IJCAI, and AAAI. He serves as an Associate Editor for IEEE TPAMI, IJCV, and ACM TOMM, and previously served for IEEE TCSVT and IEEE TMM.
\end{IEEEbiography}

\onecolumn
\appendix

\subsection{Comparison on Inception Score} \label{sec:is}

We also show the comparison on Inception Score (IS) in Fig.~\ref{fig:is}. We see that adaLN-Step1 outperforms adaLN but is inferior to adaLN-Zero in Fig.~\ref{fig:is} (a), indicating again that adding the scaling element $\alpha$ is effective in improving model performance. Also, we observe that adaLN-Mix has a marginal enhancement over adaLN-Step1, implying that the discrepancy in gradient update is not the key reason for the large disparity between adaLN-Zero and adaLN-Step1. At the same time, in Fig.~\ref{fig:is} (b), in the initial iterations when the discrepancy of gradient update occurs, we do not see any significant variation on IS, which also demonstrates that the influence of the update discrepancy is not critical.

\begin{figure}[h]
\centering\centerline{\includegraphics[width=1.0\linewidth]{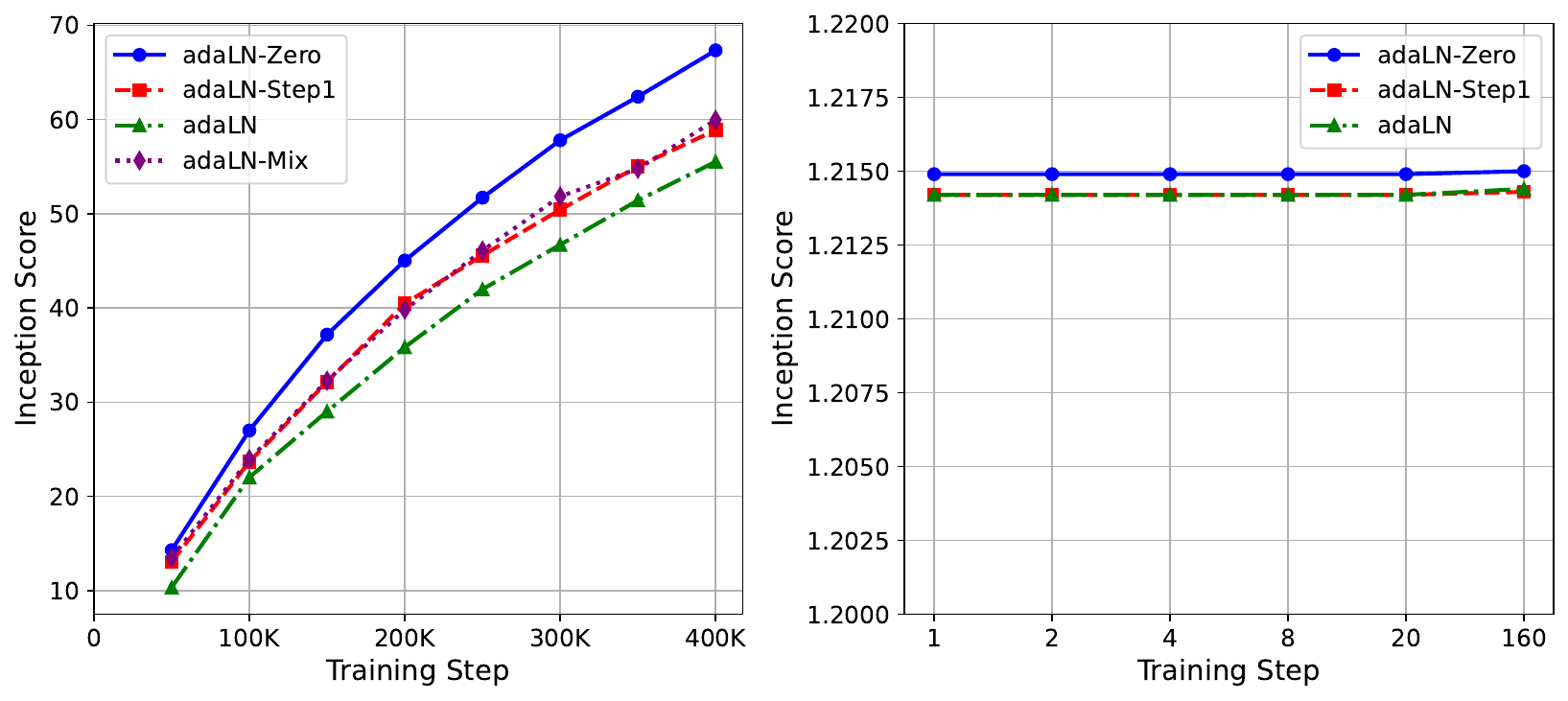}}
\hspace{1cm} (a)  \hspace{8.5cm} (b)
\caption{Comparing adaLN-Zero with adaLN as well as different initialization strategies on Inception Score (IS). We use the largest model DiT-XL/2 for all the experiments above.}
\label{fig:is}
\end{figure}

\subsection{The Structure of Squeeze-and-Excitation Module} \label{App:se-structure}
\begin{wrapfigure}{r}{0.2\textwidth}
\vskip -0.3in
    \includegraphics[width=0.2\textwidth]{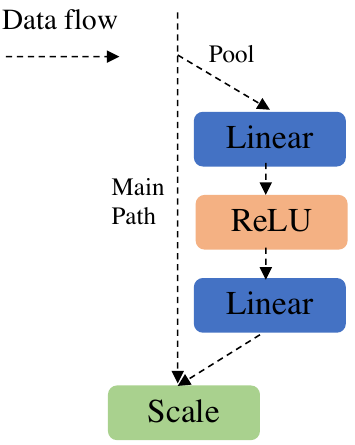} 
    \caption{The structure of SE module.}
    \label{fig:SE-structure}
\end{wrapfigure}
In Fig.~\ref{fig:SE-structure}, we illustrate the structure of Squeeze-and-Excitation (SE) module. We can see that SE module serves as a side pathway compared to the main path.

\subsection{Gradient Derivation of A simplified DiT} \label{app:gradient dereviation}
 
To calculate the loss, for simplicity, we only consider MSE loss given the target noise $\epsilon$ sampled from $N(0, I)$ and formulate $\mathcal{L}$ as $\mathcal{L} = \frac{1}{C}\sum_{i=1}^{m} \sum_{j=1}^{n} (\bar{\epsilon}_{ij} - \epsilon_{ij})^2$, where $C=m*n$ and $\epsilon_{ij}$ is the element in row $i$ and column $j$. With this formula, we can obtain $\frac{\partial \mathcal{L}}{\partial \bar{\epsilon}_{ij}} = \frac{2}{C}(\bar{\epsilon}_{ij} - \epsilon_{ij})$. Hence, we obtain the following general formula:
\begin{equation} \label{eq:loss gradient}
 \frac{\partial \mathcal{L}}{\partial \bar{\epsilon}} = \frac{2}{C}(\bar{\epsilon} - \epsilon) \,.
\end{equation}
Further, built on Eq.~\ref{eq:loss gradient}, we can also derive the gradient of $W_{f}$, $W_{ffm}$, $W_{att}$, and $W_{pat}$, \ie, $\frac{\partial \mathcal{L}}{\partial W_{f}}$, $\frac{\partial \mathcal{L}}{\partial W_{ffm}}$, $\frac{\partial \mathcal{L}}{\partial W_{att}}$, $\frac{\partial \mathcal{L}}{\partial W_{pat}}$, respectively. Before we present these formulas, we first introduce a substitution to ease our calculation:
\begin{align} 
    \bar{\epsilon} &= x_{f} * W_f \label{eq:Wf} \\ 
       &= \left\{[(x_{m_2} * W_{ffm}) \odot \alpha_2 + x_{out_{1}}] \odot (1+\gamma_f) + \beta_f\right\} * W_f  \label{eq:Wffm} \\
       &= \left\{\begin{aligned}
            &[((((x_{m_1} * W_{att}) \odot \alpha_1 + x_p) \odot (1+\gamma_2) + \beta_2) * W_{ffm}) \odot \alpha_2 \\
         & + (x_{m_1} * W_{att}) \odot \alpha_1 + x_p] \odot (1+\gamma_f) + \beta_f 
            \end{aligned}\right\} * W_f \label{eq:Watt} \\  
        &= \left\{\begin{aligned}
            &[((((((x * W_{pat}) \odot (1+\gamma_1) + \beta_1) * W_{att}) \odot \alpha_1 + (x * W_{pat})) \\  &  \odot (1+\gamma_2) + \beta_2) * W_{ffm}) \odot \alpha_2 + (((x * W_{pat})\odot (1+\gamma_1) 
            \\ &  + \beta_1) * W_{att}) \odot \alpha_1 + (x * W_{pat})] \odot (1+\gamma_f) + \beta_f 
            \end{aligned}\right\} * W_f  \label{eq:Wpat} \,.
\end{align}

With this substitution, we can derive $\frac{\partial \mathcal{L}}{\partial W_{f}}$ from Eq.~\ref{eq:Wf}. Similarly, $\frac{\partial \mathcal{L}}{\partial W_{ffm}}$, $\frac{\partial \mathcal{L}}{\partial W_{att}}$, and $\frac{\partial \mathcal{L}}{\partial W_{pat}}$ can be derived from Eqs.~\ref{eq:Wffm}, ~\ref{eq:Watt}, and ~\ref{eq:Wpat}, respectively. Thus, with the help of~\cite{laue2018computing}~\footnote{\url{https://www.matrixcalculus.org/}}, we derive the corresponding gradients as follows:
\begin{equation} \label{eq:}
\small
\frac{\partial \mathcal{L}}{\partial W_{f}} =  x^{\top}_{f} * \frac{\partial \mathcal{L}}{\partial \bar{\epsilon}} * I^\top = x^\top_{f} * \frac{2}{C} (\bar{\epsilon} - \epsilon) \,,
\end{equation}
\begin{equation}
   \frac{\partial \mathcal{L}}{\partial W_{ffm}} = x^\top_{m_2} * (( \frac{2}{C} (\bar{\epsilon} - \epsilon) * W^\top_{f} ) \odot (1+\gamma_f) \odot \alpha_2 ) \,,
\end{equation}
\begin{equation}
      \frac{\partial \mathcal{L}}{\partial W_{att}} =  x_{m_1}^\top \cdot ((( T_{0} \odot \alpha_2)\cdot W_{ffm}^\top )\odot  (1+\gamma_2) \odot \alpha_1)+  x_{m_1}^\top \cdot (T_{0} \odot \alpha_1) \,,
\end{equation}
where 
\begin{equation}
    T_{0} = (\frac{2}{C} (\bar{\epsilon} - \epsilon)*  W_{f}^\top )\odot (1+\gamma_f) \,,
\end{equation}
and 
\begin{equation}
      \frac{\partial \mathcal{L}}{\partial W_{pat}} = x^\top \cdot (((T_{2}\odot \alpha_1)\cdot W_{att}^\top )\odot (1+ \gamma_1))+ x^\top \cdot T_{2}+ x^\top \cdot (((T_{1}\odot \alpha_1 )\cdot W_{att}^\top )\odot (1+ \gamma_1))+ x^\top \cdot T_{1}
\end{equation}
where $T_{1}$ is
\begin{equation}
    T_{1} = (\frac{2}{C}(\bar{\epsilon}-\epsilon) * W_{f}^\top )\odot (1 + \gamma_f) \,,
\end{equation}
$T_{2}$ is
\begin{equation}
    T_{2} = ((T_{1}\odot \alpha_2) * W_{ffm}^\top )\odot (1+ \gamma_2) \,.
\end{equation}

Besides these parameters that are directly involved in the calculations above, we also need to determine how $\gamma_f$, $\beta_f$, $\gamma_2$, $\beta_2$, $\alpha_2$, $\gamma_1$, $\beta_1$, and $\alpha_1$ are updated, as they affect both the gradients of the above parameters and the output prediction. Hence, we provide the corresponding gradients below, respectively (omitting the bias term for simplicity):

\begin{equation} \label{eq:}
\frac{\partial \mathcal{L}}{\partial W_{\gamma_f}} =  (c \odot {\sigma(c)})^\top * ((\frac{2}{C}(\bar{\epsilon}-\epsilon)* W_{f}^\top )\odot x_{out_{2}}) \,,
\end{equation}
\begin{equation} \label{eq:}
\frac{\partial \mathcal{L}}{\partial W_{\beta_f}} = (c \odot {\sigma(c)})^\top * \frac{2}{C}(\bar{\epsilon}-\epsilon)* W_{f}^\top   \,,
\end{equation}
\begin{equation} \label{eq:}
\frac{\partial \mathcal{L}}{\partial W_{\alpha_2}} = (c \odot {\sigma(c)})^\top * (T_{1}\odot x_{ffm})
   \,,
\end{equation}
\begin{equation} \label{eq:}
\frac{\partial \mathcal{L}}{\partial W_{\gamma_2}} = (c \odot {\sigma(c)})^\top * (((T_{1} \odot \alpha_2 )* W_{ffm}^\top )\odot x_{out_{1}})
   \,,
\end{equation}
\begin{equation} \label{eq:}
\frac{\partial \mathcal{L}}{\partial W_{\beta_2}} = (c \odot {\sigma(c)})^\top * (T_{1}\odot \alpha_2)* W_{ffm}^\top 
   \,,
\end{equation}
\begin{equation}
      \frac{\partial \mathcal{L}}{\partial W_{\alpha_1}} =  (c \odot {\sigma(c)})^\top* ( T_{2} \odot T_{3})+ (c \odot {\sigma(c)})^\top* (T_{1}\odot T_{3}) \,,
\end{equation}
Where $T_{3}$ 
\begin{equation}
    T_{3} = ((\beta_1 + (x* W_{pat})\odot (1+\gamma_1) )* W_{att}) \,.
\end{equation}
\begin{equation}
      \frac{\partial \mathcal{L}}{\partial W_{\gamma_1}} = (c \odot {\sigma(c)})^\top* (((T_{2}\odot \alpha_1)* W_{att}^\top )\odot x_{p})+ (c \odot {\sigma(c)})^\top * (((T_{1}\odot \alpha_1)* W_{att}^\top )\odot x_{p})) \,,
\end{equation}
and
\begin{equation}
      \frac{\partial \mathcal{L}}{\partial W_{\beta_1}} = (c \odot {\sigma(c)})^\top* (T_{2}\odot \alpha_1)* W_{att}^\top +(c \odot {\sigma(c)})^\top* (T_{1}\odot \alpha_1)* W_{att}^\top \,,
\end{equation}
Where $c$ is the condition input and $\sigma(\cdot)$ is $sigmoid$ function.

\subsection{Value Distribution of the Whole $W_{\beta}$ in DiT Blocks} \label{app: all beta}
We present the value distributions of the whole $W_{\beta}$ of DiT-XL/2 using adaLN-Zero and adaLN-Mix trained for 400K iterations in Fig.~\ref{fig:beta}. Similar to $W_{\gamma}$, $W_{\beta}$ quickly exhibits a pattern similar to that of $W_{\alpha}$ at a very early stage regardless of whether it is adaLN-Zero or adaLN-Mix.

\begin{figure}[h]
\centering\centerline{\includegraphics[width=1.0\linewidth]{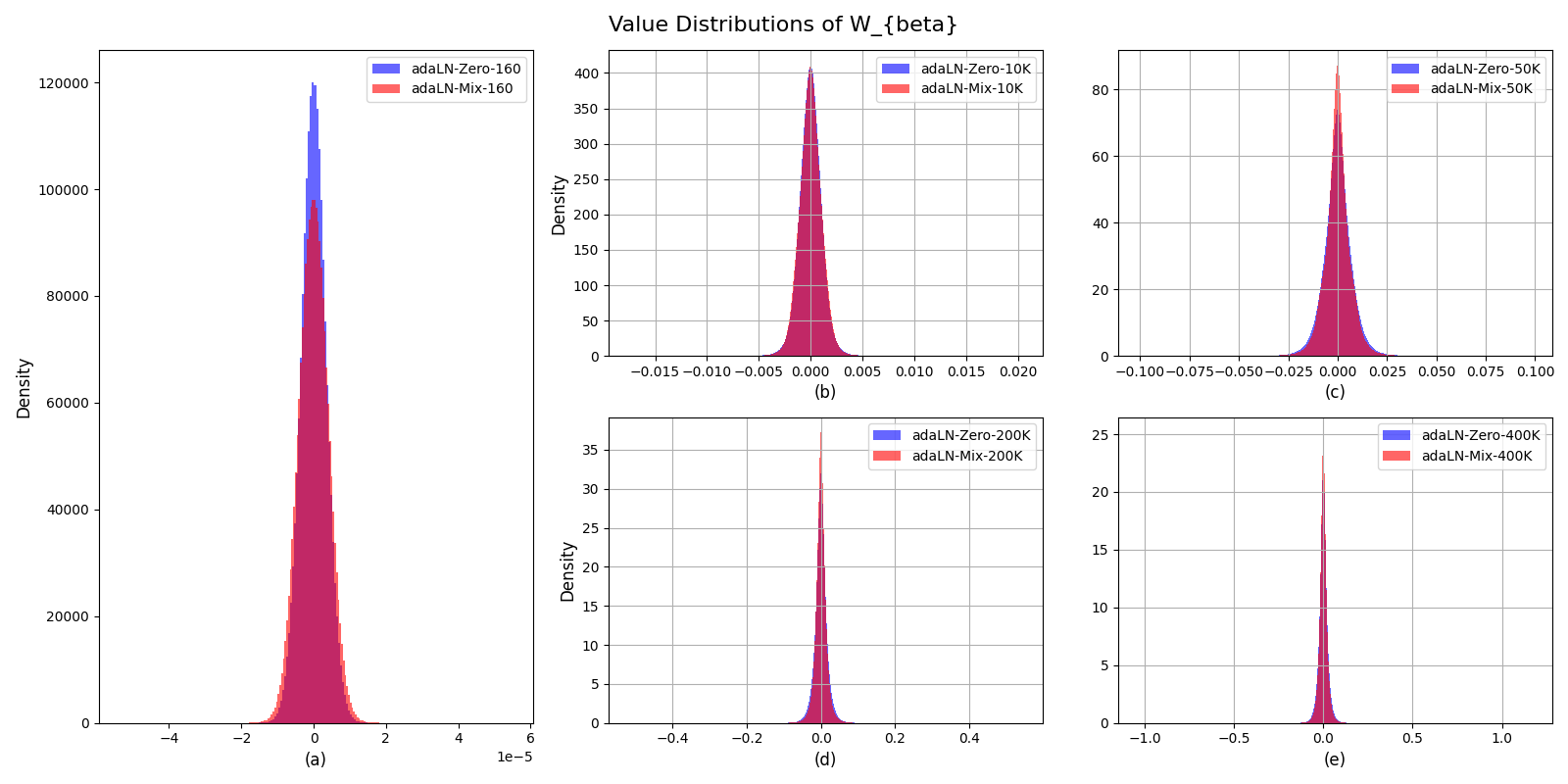}}
\caption{Value distributions of the whole $W_{\beta}$ in DiT blocks during the training process.}
\label{fig:beta}
\end{figure}

\subsection{Value Distributions of $W_{\gamma}^{L}$ and $W_{\beta}^{L}$ in Different Blocks} \label{app: different gamma and beta}

We also present the value distributions of $W_{\gamma}^{L}$ and $W_{\beta}^{L}$ in different blocks of DiT-XL/2 using adaLN-Zero trained at a very early stage (for 10K iterations). Fig.~\ref{fig:gamma-block} and Fig.~\ref{fig:beta-block} illustrate the results of $W_{\gamma}^{L}$ and $W_{\beta}^{L}$, respectively. One can see that, basically, the distributions of $W_{\gamma}^{L}$ and $W_{\beta}^{L}$ in each block share a similar pattern to their global ones as well as that of $W_{\alpha}$. Moreover, similar to $W_{\alpha}^{L}$, the peak value and bottom width of $W_{\gamma}^{L}$ and $W_{\beta}^{L}$ vary across blocks and exhibit different std, reflecting the update preference of each block. Built on this observation, this motivates us to initialize $W_{\alpha}$, $W_{\gamma}$, and $W_{\beta}$ with a more sophisticated initialization strategy. More details are in \textcolor{red}{App.~\ref{app:block wise initialization}}.

\begin{figure}[h]
\vskip -0.2in
\centering\centerline{\includegraphics[width=1.0\linewidth]{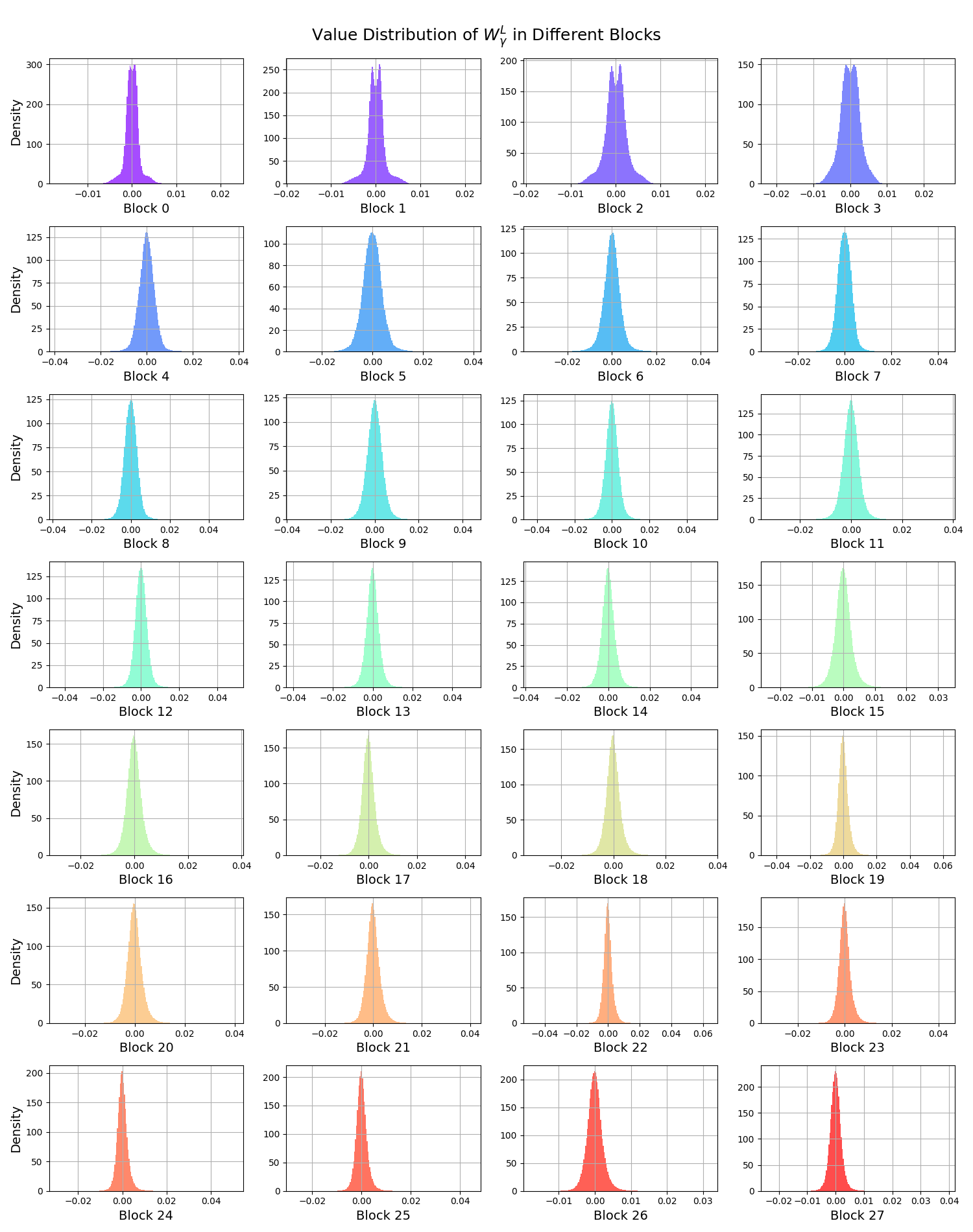}}
\caption{Value distributions of $W_{\gamma}^{L}$ in different blocks.}
\label{fig:gamma-block}
\vskip -0.1in
\end{figure}

\begin{figure}[h]
\vskip -0.2in
\centering\centerline{\includegraphics[width=1.0\linewidth]{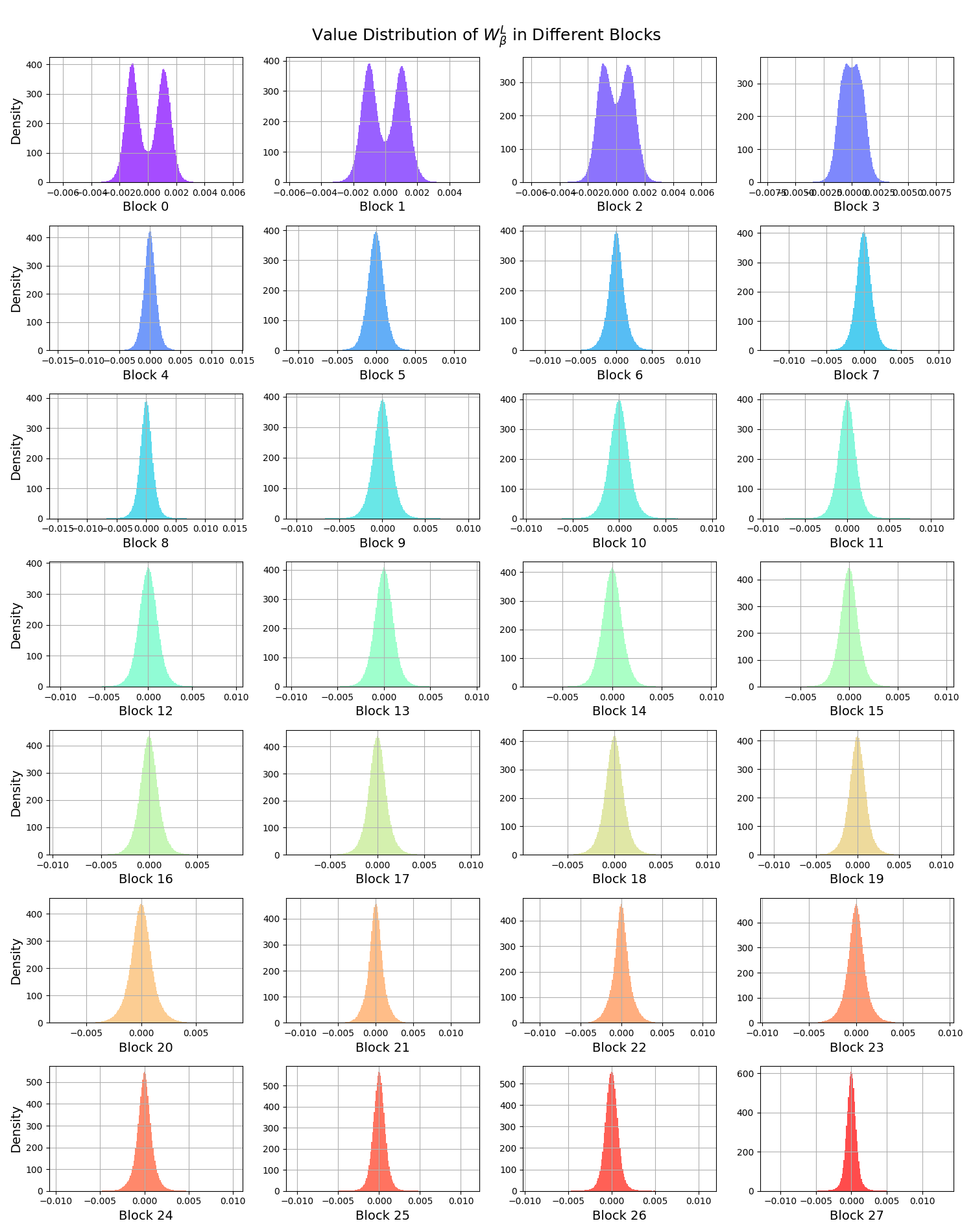}}
\caption{Value distributions of $W_{\beta}^{L}$ in different blocks.}
\label{fig:beta-block}
\vskip -0.2in
\end{figure}

\clearpage

\begin{figure}[h]
\vskip -0.3in
\centering\centerline{\includegraphics[width=1.0\linewidth]{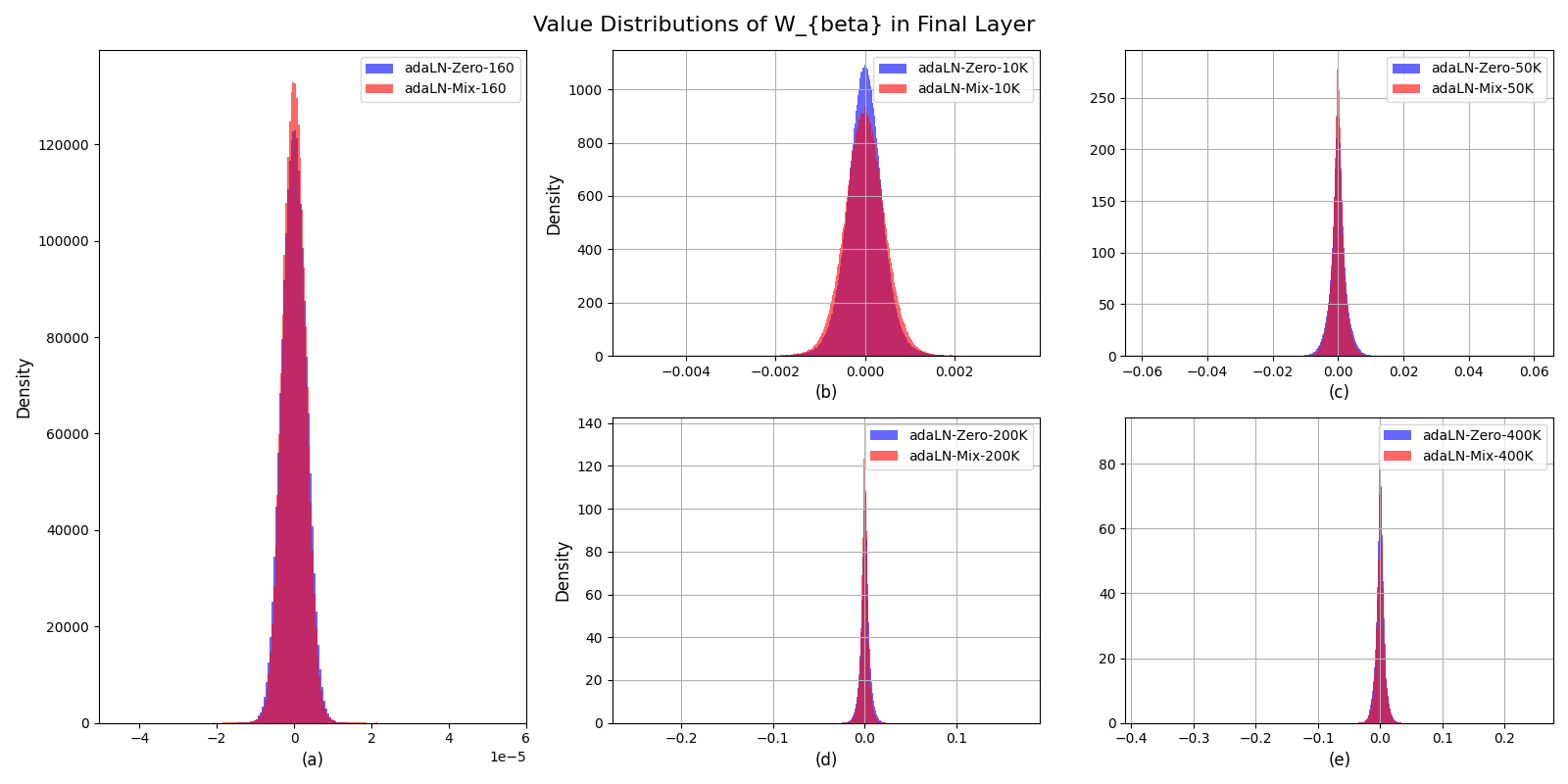}}
\vskip -0.1in
\caption{Value distributions of $W_{\beta_f}$ in FinalLayer during the training process.}
\label{fig:beta_f}
\vskip -0.1in
\end{figure}

\begin{figure}[h]
\vskip -0.2in
\centering\centerline{\includegraphics[width=1.0\linewidth]{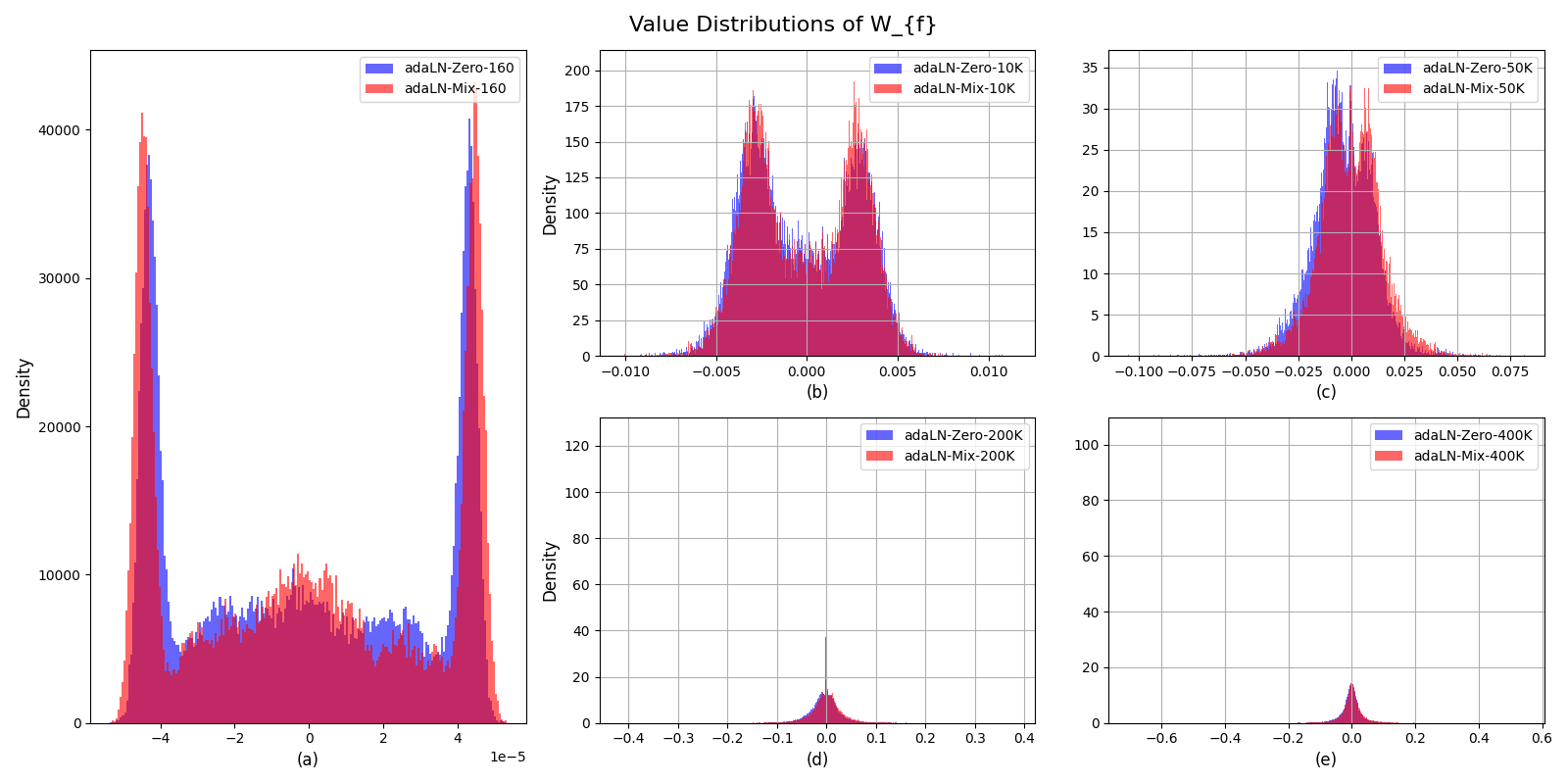}}
\vskip -0.1in
\caption{Value distributions of $W_{f}$  in FinalLayer during the training process.}
\label{fig:w_f}
\end{figure}

\subsection{Value Distributions of $W_{\beta_f}$ and $W_{f}$} \label{app: W_f}

\begin{wraptable}{r}{4cm}
\centering
\vskip -0.6in
\caption{A grid search of std for $W_{\beta_f}$. $0$: AdaLN-Gaussian-v1}
\small
\begin{tabular}{l c c}
	\toprule
        Std  & FID  & IS
   \\
	\midrule
 0 &  76.21 & 15.01  \\
\midrule
 2e-4 & 78.22 & 14.53  \\
 5e-4 & 82.05 & 13.78 \\
 1e-3 & 80.43  & 14.03 \\
 2e-3 & 77.45 & 14.74 \\
 3e-3 & 77.09 & 14.84 \\
 4e-3 & 78.47 & 14.39 \\
\bottomrule
\end{tabular}
\label{tab:w_beat_f_std}
\end{wraptable}

Fig.~\ref{fig:beta_f} and Fig.~\ref{fig:w_f} illustrate the variations of value distribution of $W_{\beta_f}$ and $W_f$ under different training time. We can see that $W_{\beta_f}$ and $W_f$ present completely different variation tendencies. Even though, we also notice that $W_{\beta_f}$ shares a similar pattern to $W_{\alpha}$ at a very early stage regardless of whether it is adaLN-Zero or adaLN-Mix. This inspires us to explore whether initializing $W_{\beta_f}$ together with $W_{\alpha}$, $W_{\gamma}$, and $W_{\beta}$ could further accelerate training. Based on the setting $std(1e-3, 1e-3, 1e-3)$ for $W_{\alpha}$, $W_{\gamma}$, and $W_{\beta}$, we perform a grid search of std for $W_{\beta_f}$ as shown in Tab.~\ref{tab:w_beat_f_std}. It appears that initializing $W_{\beta_f}$ with a wide range of std values does not enhance the model's performance. In light of these results, we do not consider initializing $W_{\beta_f}$ with Gaussian and keep its original zero-initialization strategy for all the experiments.

\subsection{Value Distributions of Non-zero-initialized DiT Modules}~\label{App:More DiT Modules}
We visualize the value distribution of non-zero-initialized DiT modules including Attention and Mlp in DiT blocks, and PatchEmbed as shown in Fig.~\ref{fig:attention}, Fig.~\ref{fig:mlp}, and Fig.~\ref{fig:patchembed}, respectively. Though they are all initialized with Xavier uniform initialization in DiT, the weight distributions in both Attention and MLP gradually transition to a Gaussian-like distribution while PatchEmbed does not. 
We also visualize the value distribution of LabelEmbedder and TimestepEmbedder in Fig.~\ref{fig:label-time}. We see that after normal initialization done in DiT, their weight distributions consistently show a Gaussian-like distribution. Naturally, we can consider Gaussian initializations for these modules as well except PatchEmbed to accelerate training. For example, we could uniformly use Gaussian initialization for Attention and Mlp in DiT Block. We set the mean to $0$ and use several choices for std such as $0.001$, $0.01$, $0.02$, $0.03$, and $0.04$. We use DiT-XL-2 and train for 50K steps for simplicity. The results are shown in Tab.~\ref{tab:other-parts}. 
We see that the performance is inferior to the default initialization. We further consider to leverage different std for attention and MLP since the distribution widths of attention and MLP weights are different as shown in Fig~\ref{fig:attention} and Fig.~\ref{fig:mlp}.  Specifically, based on the results in Tab.~\ref{tab:other-parts}, we set std to 0.02 for attention and set std to 0.01 for MLP. This new setting produces 74.40 for FID and outperforms the default setting (76.21 FID), demonstrating the generalization of the improvements observed in adaLN-Gaussian to other modules. To further unleash the potential, more precise hyperparameter tuning may be needed for these modules, which we leave as future work.

\begin{figure}[t]
\vskip -0.3in
\centering\centerline{\includegraphics[width=1.0\linewidth]{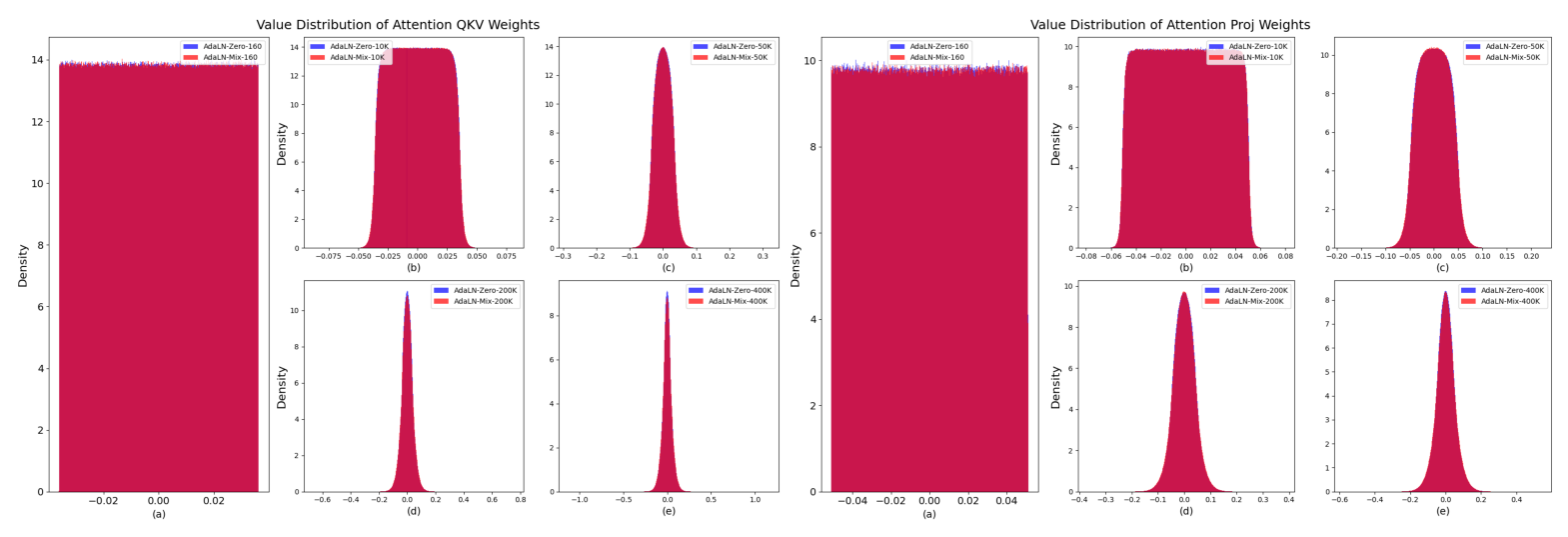}}
\vskip -0.1in
\caption{Value distributions of Attention module including qkv and proj during training process.}
\label{fig:attention}
\vskip -0.05in
\end{figure}

\begin{figure}[t]
\vskip -0.1in
\centering\centerline{\includegraphics[width=1.0\linewidth]{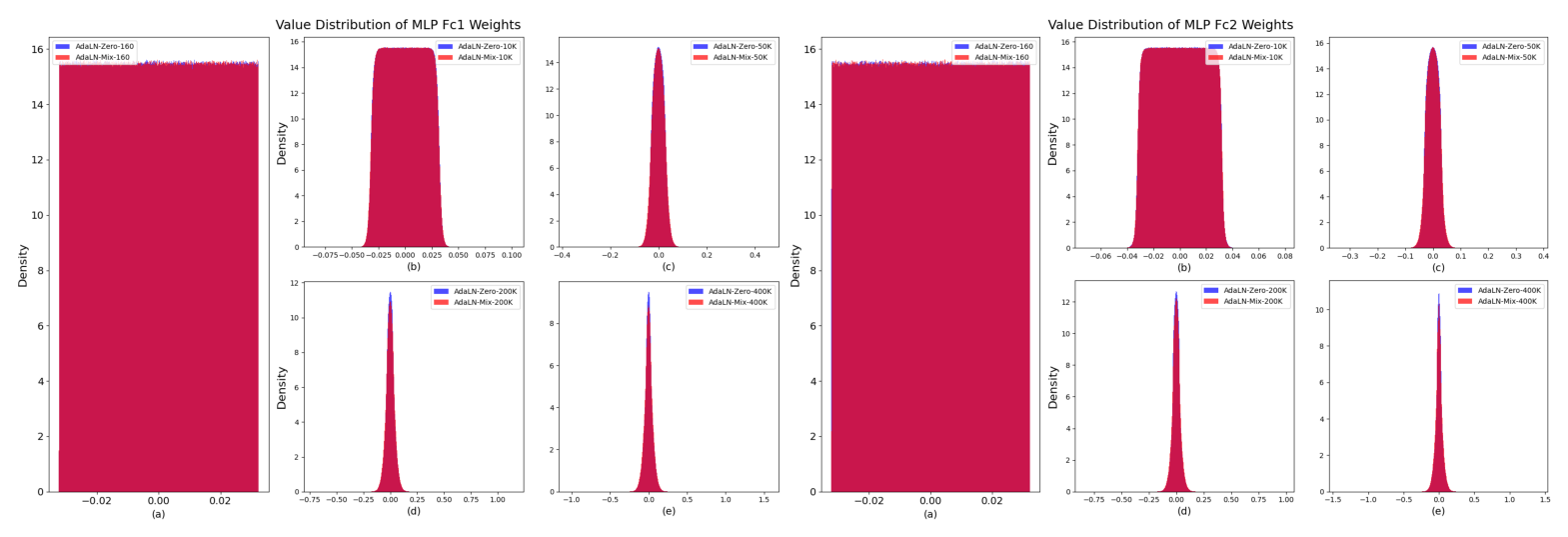}}
\vskip -0.1in
\caption{Value distributions of Mlp module including fc1 and fc2 during training process.}
\label{fig:mlp}
\vskip -0.2in
\end{figure}

\begin{figure}[h]
\vskip -0.1in
\centering\centerline{\includegraphics[width=1.0\linewidth]{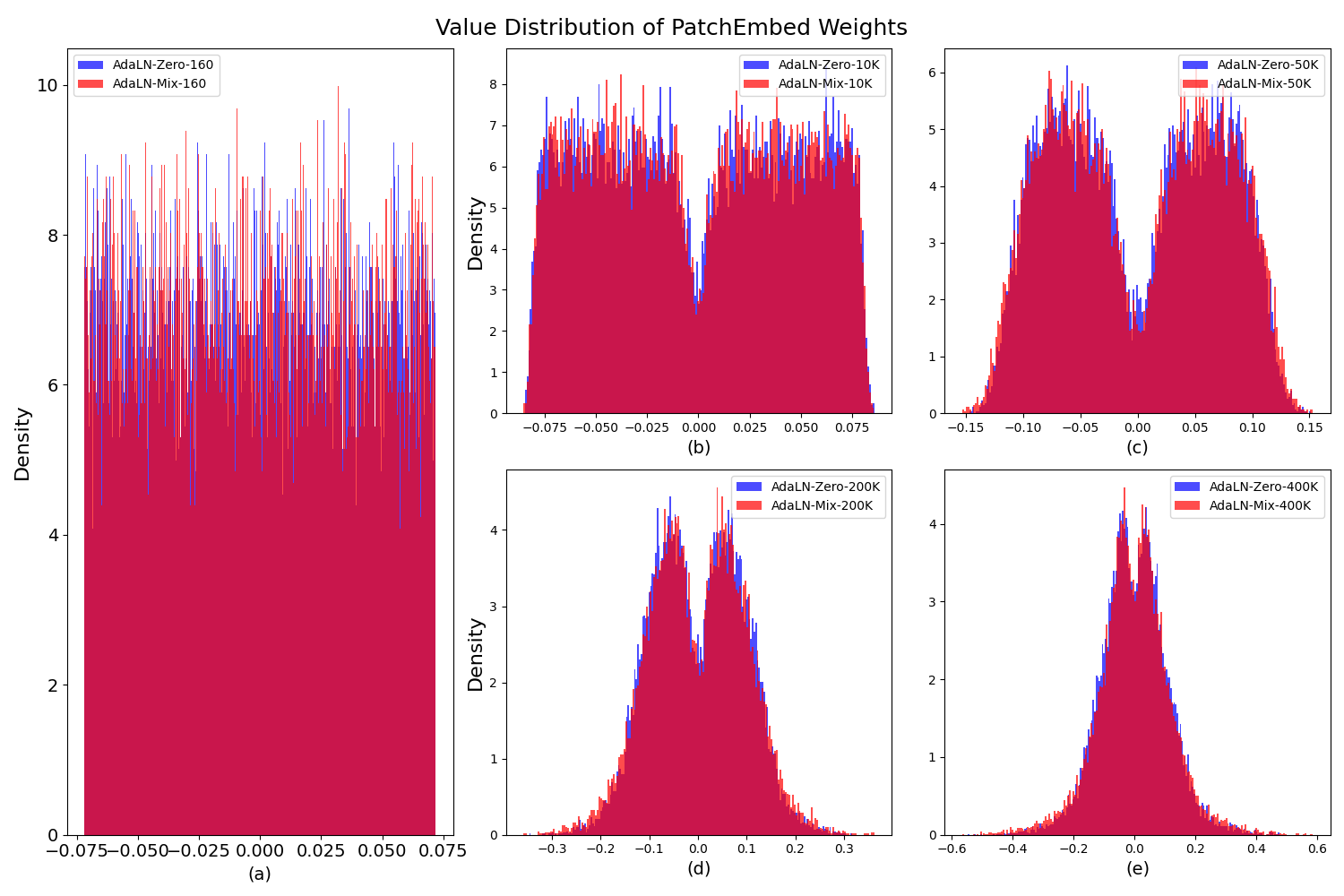}}
\vskip -0.1in
\caption{Value distributions of PatchEmbed during training process.}
\label{fig:patchembed}
\vskip -0.1in
\end{figure}

\begin{figure}[h]
\centering\centerline{\includegraphics[width=1.0\linewidth]{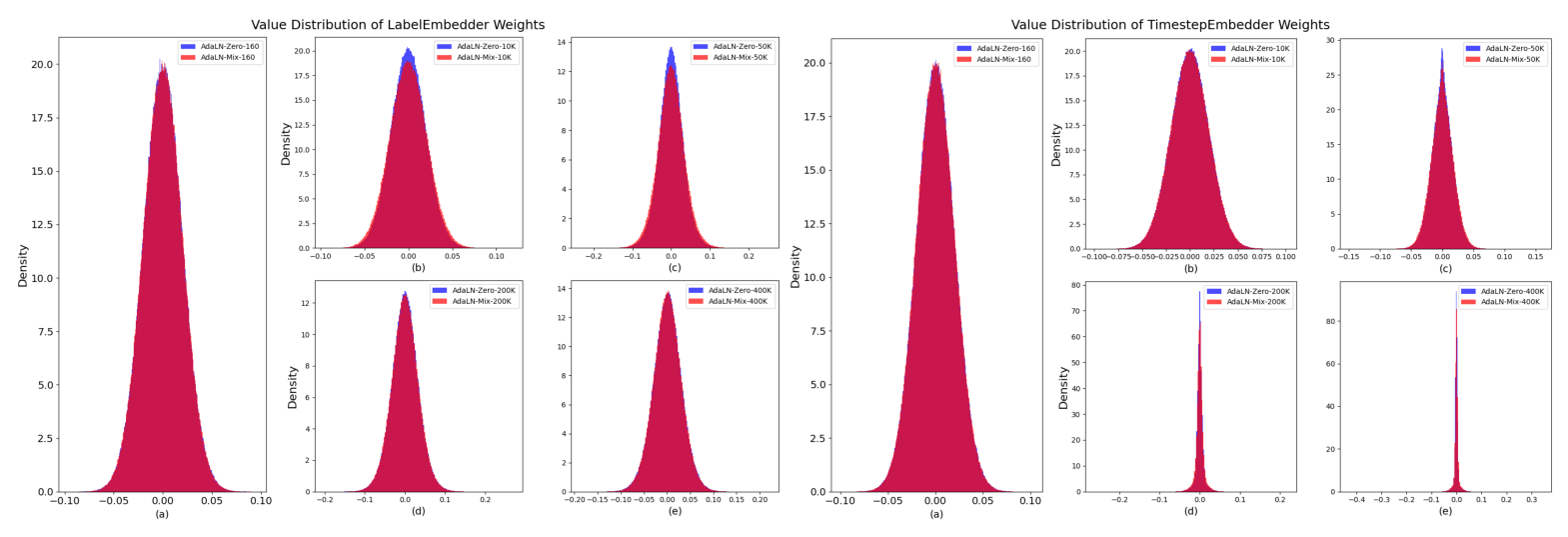}}
\caption{Value distributions of LabelEmbedder and TimestepEmbedder during training process.}
\label{fig:label-time}
\end{figure}

\setlength{\tabcolsep}{0.6cm}{\begin{table}[h]
		\caption{Different Gaussian std initialization choices  for Attention and Mlp in DiT Block.} 
		\begin{center}
        \small
			\begin{tabular}{l c c c c c c c}
				\toprule
				   Std & Default & 0.001 & 0.01 & 0.02 & 0.03 & 0.04 & (0.01, 0.02)\\ 
				\midrule
				FID & 76.21 & 92.09 &	85.28 &	80.89 & 91.21 & 98.50 & 74.40  \\
				\bottomrule
			\end{tabular}
		\end{center}
            \vskip -0.1in
		\label{tab:other-parts}
  \vskip -0.15in
\end{table}}

\subsection{Value Distributions of zero-convolution in ControlNet}~\label{App:zero-convolution in ControlNet}
Besides adaLN-Zero in DiT, we also consider a similar module in ControlNet~\cite{zhang2023adding} called zero convolution. In Fig.~\ref{fig:ControlNet}, we visualize the weight distributions of four widely-used ControlNet variants including Canny, Depth, Pose, and Segmentation. Their distributions are still a Gaussian-like distribution. Hence, it is worth investigating whether Gaussian initialization is also beneficial for these modules in ControlNet. Since it is not our main focus, we leave it as future work.

\begin{figure}[h]
\centering\centerline{\includegraphics[width=1.0\linewidth]{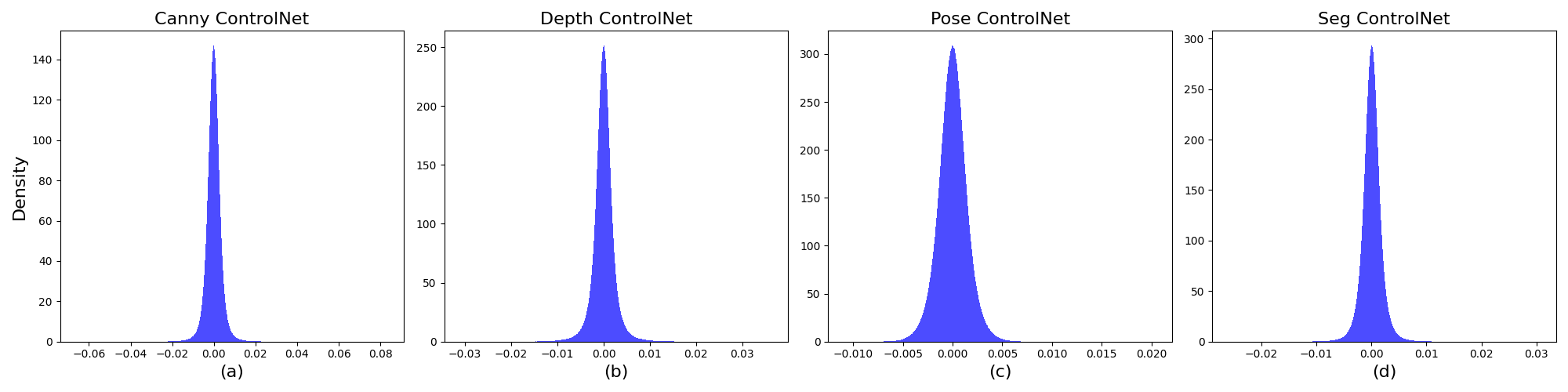}}
\caption{Weight distributions of zero convolution in four ControlNet variants.}
\label{fig:ControlNet}
\end{figure}

\subsection{Result Analysis about Different Std Choices in Gaussian Initialization}\label{App:Analysis-Different-Std}

Intuitively, since the weights of the conditional mechanisms we counted are Gaussian-like distributions, there should exist an optimal std hyperparameter when initializing these weights with Gaussian, and naturally, the values on both sides of this hyperparameter are relatively unsuitable. To some extent, the performance of Gaussian initialization with different std choices in Tab II of main paper which exhibits a U-shaped trend also proves it. To be more rigorous, we analyze this U-shaped trend by leveraging two representative settings, \ie, $std=0.0005$ and $std=0.05$, which correspond to the two ends of this U-shaped trend.

We first illustrate their weight distributions of $W_{\alpha}$ in the conditioning mechanism and compare them with that of adaLN-Zero and adaLN-Gaussian (std=0.001). The results are shown in Fig.~\ref{fig:different-std}. We find that a large std$=0.05$ presents a less compact distribution and exhibits a significant discrepancy in distribution shape compared to the rest settings. \textit{This result indicates that a large std may be incompatible with other parameters, resulting in a slow speed of convergence and a poor performance.} Moreover, we consider this a step further. Theoretically, if we further increase the std value, it would become close to the default initialization in adaLN-Step1 (xavier\_uniform) while the performance of adaLN-Step1 is also bad.

For a small std$=0.0005$, it can be seen that the distribution of $W_{\alpha}$ is quite similar to that of adaLN-Zero and adaLN-Gaussian ($std=0.001$). However, there still exists a slight discrepancy. To make this discrepancy clearer, we average the absolute values of the differences between each element in $W_{\alpha}$ corresponding to $std=0.0005$ and adaLN-Zero, and $std=0.0005$ and adaLN-Gaussian. The element-wise averaged results are 0.0121 and 0.0124, respectively. \textit{By comparing the results (0.0121 $<$ 0.0124), it is shown that small std leads to weights relatively closer to that of zero-initialization (adaLN-Zero).} And, to some extent, the corresponding performance also proves it where std=0.0005 produces 80.68 for FID, closer to adaLN-Zero (78.99) compared to adaLN-Gaussian (76.21).

\begin{figure}[h]
\centering\centerline{\includegraphics[width=1.0\linewidth]{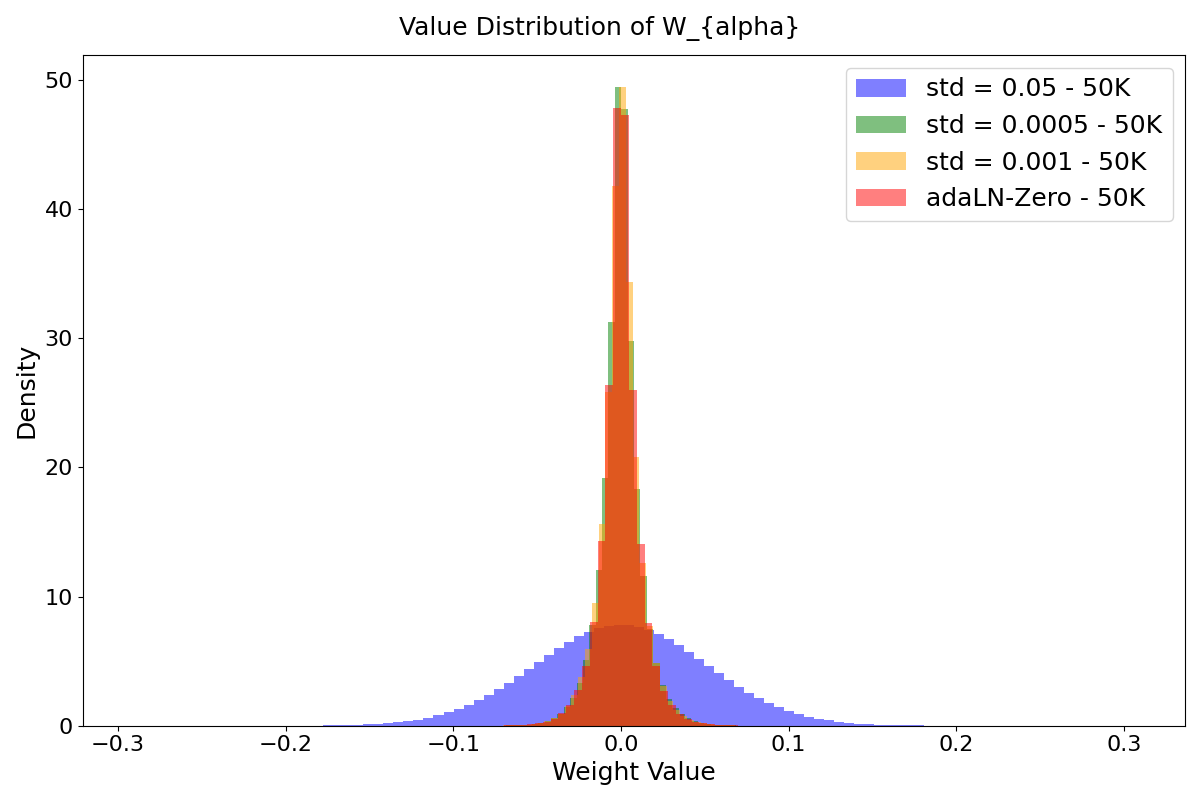}}
\caption{Value distributions of $W_{\alpha}$ with different std in Gaussian initialization.}
\label{fig:different-std}
\end{figure}

\newpage

\subsection{AdaLN-Gaussian-v2}\label{app:adaLN-Gaussian-v2}

\begin{wraptable}{r}{6.1cm}
\vskip -0.18in
\centering
\caption{Results of independent std settings for $W_{\alpha}$, $W_{\gamma}$, $W_{\beta}$. 0, 0, 0: adaLN-Zero}
\vskip -0.05in
\setlength{\tabcolsep}{8pt}
\small
\begin{tabular}{l c c}
	\toprule
        Std ($W_{\alpha}$, $W_{\gamma}$, $W_{\beta}$) & FID  & IS
   \\
   \midrule
    0, 0, 0 & 78.99 & 14.19  \\
 1e-3, 2e-3, 8e-4 & 78.22 & 14.37 \\
 1e-3, 1.2e-3, 8e-4 & 76.57 & \textbf{15.01} \\
 \rowcolor{yellow!20} 8e-4, 1.2e-3, 8e-4 & \textbf{76.12} & 14.90 \\
 8e-4, 1.2e-3, 1e-3 & 77.18 &  14.85 \\
 8e-4, 1e-3, 8e-4 & 80.31 & 14.23 \\
8e-4, 1.4e-3, 8e-4 & 77.53 & 14.54 \\
8e-4, 1.6e-3, 8e-4 & 78.24 & 14.55 \\
 8e-4, 1.6e-3, 4e-4 & 79.03 & 14.31 \\
\bottomrule
\end{tabular}
\vskip -0.1in
\label{tab:fined_std}
\end{wraptable}

We begin by considering $std(1e-3, 2e-3, 8e-4)$~\footnote{We empirically find that $N(0, 1e-3)$ closely matches the shape of the $W_{\alpha}$ distribution in Fig.4 (b) of main paper (10K iterations). Therefore, based on this observation, we begin our further refinement by estimating the std for $W_{\gamma}$ and $W_{\beta}$ with their corresponding distribution shapes in 10K iterations.}, restrict the range from 8e-4 to 2e-3, inspired by Tab.II of main paper, and perform a grid search in Tab.~\ref{tab:fined_std}. It is observed that $std(8e-4, 1.2e-3, 8e-4)$ produces the best FID. We denote this initialization as \textit{adaLN-Gaussian-v2}.

Based on adaLN-Gaussian-v2, we further explore a more sophisticated block-wise initialization. This is motivated by our observation that the peak value and bottom width of $W_{\alpha}^{L}$, $W_{\gamma}^{L}$, and $W_{\beta}^{L}$ varies across DiT blocks in Fig.6 of main paper, Fig.~\ref{fig:gamma-block}, and Fig.~\ref{fig:beta-block}, indicating that different blocks may prefer different std. In our preliminary attempt in \textcolor{red}{App.~\ref{app:block wise initialization}}, we show that block-wise initialization is inferior to the base setting in FID but outperforms the base setting in IS. This highlights the potential of block-wise initialization and requires more effort, which we leave as future work. 

Furthermore, we compare the performance of adaLN-Gaussian-v2 with adaLN-Zero and adaLN-Gaussian under longer training time as shown in Tab~\ref{tab:three-comparison}. It is seen that adaLN-Gaussian-v2 also outperforms adaLN-Zero under the same steps, further verifying the effectiveness of our strategy of Gaussian initialization on improving training efficiency. On the other hand, considering that adaLN-Gaussian achieves better results than adaLN-Gaussian-v2
and is easier to implement, we primarily use adaLN-Gaussian in Tab IV of main paper.

\setlength{\tabcolsep}{0.25cm}{\begin{table}[h]
\centering
        \caption{Comparison among adaLN-Zero, adaLN-Gaussian, and adaLN-Gaussian-v2.} 
			\small
			\begin{tabular}{l l c l c c c c c}
				\toprule
				Model & Initialization & CFG & Steps &  FID$\downarrow$  & sFID$\downarrow$ & IS$\uparrow$ & Precision$\uparrow$ & Recall$\uparrow$ \\ 
				\midrule
                    \midrule
				DiT-XL/2 &  adaLN-Zero  & 1 & 400K  & 20.02  & 6.09 & 67.34 & 63.33 & 63.06  \\
    			DiT-XL/2
				& adaLN-Gaussian & 1 & 400K  & 17.86  &  6.06 & 73.07 & 64.51 & 62.64  \\
                    DiT-XL/2 & adaLN-Gaussian-v2 & 1 & 400K  & 18.77  & 6.08 & 70.07 & 63.92 &  62.72 \\
				\bottomrule
			\end{tabular}
		\label{tab:three-comparison}
  \vskip -0.15in
\end{table}}

\subsection{A Preliminary Exploration of Block-wise Initialization}~\label{app:block wise initialization}
\begin{wrapfigure}{r}{0.4\textwidth}
    \centering
    \vskip -0.35in
    \includegraphics[width=0.38\textwidth]{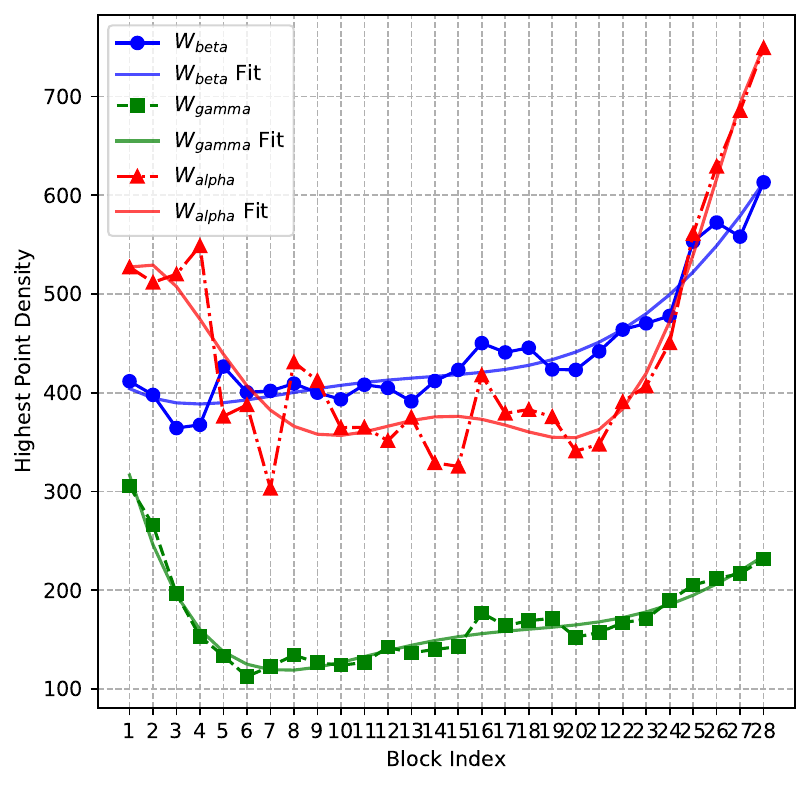}
    \vskip -0.05in
    \caption{Three polynomial functions to fit the peak values of $W_{\alpha}$, $W_{\gamma}$, and $W_{\beta}$ in all blocks.}
    \vskip 0.1in
    \label{fig:curve}
\end{wrapfigure}
We dive into every block in DiT and find that there also exist discrepancies in peak value among different $W_{\alpha}^{L}$ in Fig.~6 in main paper. $W_{\gamma}^{L}$ and 
$W_{\beta}^{L}$ also hold in Fig.~\ref{fig:gamma-block} and Fig.~\ref{fig:beta-block}. 
Generally, the greater the peak value is, the smaller the std is, motivating us to design a more sophisticated block-wise initialization strategy. Specifically, we record the peak value in all blocks for $W_{\alpha}^{L}$, $W_{\gamma}^{L}$, and $W_{\beta}^{L}$, respectively, and use three heuristic polynomial functions to fit these points as shown in Fig.~\ref{fig:curve}. For $W_{\alpha}^{L}$, we use 7th degree polynomial whose coefficients are [$-2.49635921e-6$,  $1.24680129e-4$,  $1.17149262e-3$, $-1.70585560e-1$, $3.63484494$, $-2.94971466e+1$,  $6.74700382e+1$,  $4.85897902e+2$]. For $W_{\gamma}^{L}$ and 
$W_{\beta}^{L}$, we use 5th degree polynomial. Their coefficients are [$-2.46024908e-4$,  $2.39970674e-2$, $-8.67912602e-1$,  $1.45429227e+1$, $-1.08645122e+2$,  $4.11540316e+2$] and [$-1.21059796e-4$, $1.09417700e-2$, $-3.25623123e-1$,  $4.15804173$, $-2.00345083e+1$,  $4.20334676e+2$], respectively. For $W_{\alpha}^{L}$ in L$-th$ block, we use the following formula to calculate its std value:

\begin{equation}
    \sigma^{L}_{\alpha} = 0.0008 / (Poly_{\alpha}(L)/449.9321) \,,
\end{equation}

where $Poly_{\alpha}$ is the polynomial function for $\alpha$, $0.0008$ is the base std inherited from Tab.~\ref{tab:fined_std}, and $449.9321$ is the averaged peak value across $W_{\alpha}^{L}$ in all blocks.

\setlength{\tabcolsep}{0.55cm}{\begin{table}
\centering
\caption{Results of block-wise initialization. \ding{52}: with block-wise }
\vskip 0.05in
\small
\begin{tabular}{l c c}
	\toprule
        Std ($W_{\alpha}$, $W_{\gamma}$, $W_{\beta}$) & FID  & IS
   \\
	\midrule
8e-4, 1.2e-3, 8e-4 & \textbf{76.12} & 14.90 \\
\midrule
\ding{56}, \quad \ding{52}, \quad \ding{52} & 79.28 & 14.35  \\
\ding{52},  \quad \ding{52},  \quad \ding{52} & 76.63 & \textbf{14.96} \\
\bottomrule
\end{tabular}
\label{tab:block_init}
\end{table}}

Similarly, for $W_{\gamma}^{L}$ and $W_{\beta}^{L}$, we use the following formulas to calculate their std value, respectively:
\begin{equation}
    \sigma^{L}_{\gamma} = 0.0012 / (Poly_{\gamma}(L)/444.8248) \,,
\end{equation}
\begin{equation}
    \sigma^{L}_{\beta} = 0.0008 / (Poly_{\beta}(L)/175.0044) \,.
\end{equation}

We first consider employing block-wise initialization for $W_{\gamma}^{L}$ and 
$W_{\beta}^{L}$ since they are well fitted and use $0.0008$ for $W_{\alpha}^{L}$ by default. Afterward, we initialize them all in a block-wise manner. As shown in Tab.~\ref{tab:block_init}, block-wise initialization is inferior to the base setting in FID50K but outperforms the base setting in IS. We leave more exploration as future work.

\subsection{More Experiments on Effectiveness}\label{App:more-experiments}

To further show the effectiveness of adaLN-Gaussian on other datasets, we add more experiments on three additional datasets including Tinyimagenet~\cite{le2015tiny}, AFHQ~\cite{choi2020stargan}, and CelebA-HQ~\cite{karras2018progressive} using the best-performing DiT-XL/2 with 50K training steps while keeping all training settings. We report all the FID results in Tab.~\ref{tab:more-experiments}. These results show that adaLN-Gaussian consistently outperforms adaLN-Zero under the same steps, effectively demonstrating the generalization of our method on improving training efficiency.

\setlength{\tabcolsep}{0.4cm}{\begin{table}[h]
		\centering
		\caption{Comparisons between adaLN-Zero and adaLN-Gaussian on another three datasets including Tinyimagenet, AFHQ and CelebA-HQ.  We set CFG=1.}
			\small
			\begin{tabular}{l c  c c }
				\toprule
				   & Tiny ImageNet & AFHQ & CelebA-HQ  \\ 
				\midrule
                    \midrule
				adaLN-Zero & 37.11   & 13.52 & 8.01   \\
 adaLN-Gaussian & 36.07 &	12.58 & 7.54  \\
				\bottomrule
			\end{tabular}
		\label{tab:more-experiments}
\end{table}}

\end{document}